\documentclass[sigconf]{acmart}%
\AtBeginDocument{%
  \providecommand\BibTeX{{%
    \normalfont B\kern-0.5em{\scshape i\kern-0.25em b}\kern-0.8em\TeX}}}
\usepackage{url}
\usepackage{amsmath}
\usepackage{array}
\usepackage{graphicx}
\usepackage{amsfonts}
\usepackage{multirow}

\usepackage{threeparttable}
\usepackage{amsmath,amsbsy}
\usepackage{bm}

\usepackage{colortbl,multirow}
\usepackage{color}
\usepackage{enumitem}
\setlist[itemize]{leftmargin=*}

\setcopyright{rightsretained} 

\begin{document}
	\fancyhead{}
\title{Deconfounded Video Moment Retrieval with Causal Intervention}
\author{Xun~Yang$^{1,2}$, Fuli Feng$^{1,2}$, Wei~Ji$^{1,2}$, Meng~Wang$^3$, Tat-Seng Chua$^{1,2}$}
\def \authors{Xun~Yang, Fuli Feng, Wei~Ji, Meng~Wang, Tat-Seng Chua}
\affiliation{%
	\institution{$^1$Sea-NExT Joint Lab
		\country{Singapore}}
}
\affiliation{%
  \institution{$^2$School of Computing, National University of Singapore
  \country{Singapore}}
}
\affiliation{%
	\institution{$^3$School of Computer Science and Information Engineering, Hefei University of Technology
		\country{China}}
}
\email{{hfutyangxun,fulifeng93}@gmail.com; wangmeng@hfut.edu.cn;{jiwei,dcscts}@nus.edu.sg}
\begin{abstract}
  We tackle the task of video moment retrieval (VMR), which aims to localize a specific moment in a video according to a textual query. Existing methods primarily model the matching relationship between query and moment by complex cross-modal interactions. Despite their effectiveness, current models mostly exploit dataset biases while ignoring the video content, thus leading to poor generalizability. We argue that the issue is caused by the hidden confounder in VMR, \textit{i.e., temporal location of moments}, that spuriously correlates the model input and prediction. How to design robust matching models against the temporal location biases is crucial  but, as far as we know, has not been studied yet for VMR.
  
  To fill the research gap, we propose a causality-inspired VMR framework that builds structural causal model to capture the true effect of query and video content on the prediction. Specifically, we develop a Deconfounded Cross-modal Matching (DCM) method to remove the confounding effects of moment location. It first disentangles moment representation to infer the core feature of visual content, and then applies causal intervention on the disentangled multimodal input based on backdoor adjustment, which forces the model to fairly incorporate each possible location of the target into consideration. Extensive experiments clearly show that our approach can achieve significant improvement over the state-of-the-art methods in terms of both accuracy and generalization (Codes: \color{blue}{\url{https://github.com/Xun-Yang/Causal_Video_Moment_Retrieval}}). 
\end{abstract}


\copyrightyear{2021} 
\acmYear{2021} 
\acmConference[SIGIR '21]{Proceedings of the 44th International ACM SIGIR Conference on Research and Development in Information Retrieval}{July 11--15, 2021}{Virtual Event, Canada}
\acmBooktitle{Proceedings of the 44th International ACM SIGIR Conference on Research and Development in Information Retrieval (SIGIR '21), July 11--15, 2021, Virtual Event, Canada}\acmDOI{10.1145/3404835.3462823}
\acmISBN{978-1-4503-8037-9/21/07}

\maketitle

\section{Introduction}\label{introd}
\begin{figure}[t]
	\centering
	\includegraphics[width=3.1in]{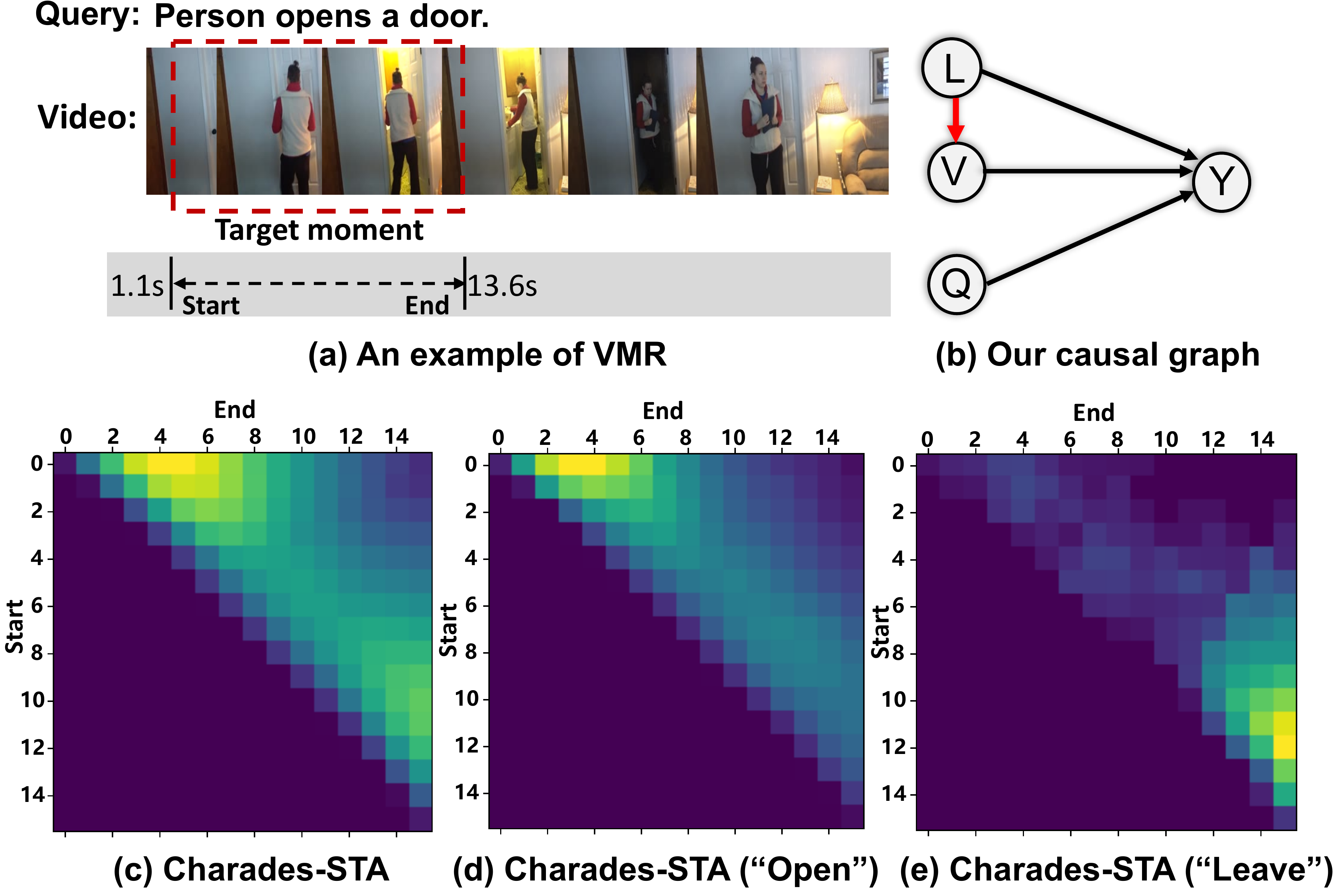}
	\vspace{-0.12in}
	\caption{(a) An example of VMR;  (b) Causal graph: $Q$ (query), $V$ (video moment), $L$ (moment location), and $Y$ (prediction); The bottom three plots depict the temporal location distributions of moment annotations in Charades-STA based on the 2D structure of moments~\cite{zhang2019learning}: (c) Distribution of all moments; (d) and (e) describe the distributions of query-specific moments (\textit{i.e.}, including verbs like "\textit{open}" or "\textit{leave}").}
	\vspace{-0.24in}
	\label{fig1}
\end{figure}

Text-based visual retrieval is a fundamental problem in multimedia information retrieval~\cite{hong2015learning,nie2012harvesting,yang2020tree,dong2018predicting,dong2021dual} which has achieved significant progress. 
Recently, Video Moment Retrieval (VMR) has emerged as a new task \cite{liu2018attentive,gao2017tall,hendricks2018localizing,anne2017localizing}
 and received increasing attention \cite{liu2018cross,chen2018temporally}. 
 It aims to retrieve a specific moment in a video, according to a textual query. As shown in Figure \ref{fig1} (a), given a query "\textit{Person opens a door}", its goal is to predict the temporal location of the target moment.\footnote{Through this paper, unless otherwise stated, we use the term "\textbf{location}" to depict the "\textbf{temporal location}" of a video moment, rather than "\textit{physical location}."} This task is quite challenging, since it requires effective modeling of multimodal content and cross-modal relationship. The past few years have witnessed the notable development of VMR, largely driven by public benchmark datasets, \textit{e.g.}, {Charades-STA}~\cite{gao2017tall}. 
Most existing efforts first generate sufficient and diverse moment candidates, and then rank them based on their matching scores with the query. Benefiting from effective modeling of temporal contexts and cross-modal interactions, state-of-the-art (SOTA) models \cite{zhang2019learning,NIPS2019_8344,mun2020LGI} have reported significant performance on public datasets. However, does the reported results of  SOTA models truly reflect the progress of VMR? The answer seems to be "\textit{No}". 

Recent studies \cite{otani2020uncovering} found empirical evidences that VMR models mostly exploit the temporal location biases in datasets, rather than learning the cross-modal matching. The true effect from the video content may be largely overlooked in the prediction. Such biased models have poor generalizability and are not effective for real-world application. We argue that the critical issue is rooted from two types of temporal location biases in dataset annotations: 
\begin{itemize}
\item  \textit{Long-tailed annotations}. As depicted in Figure \ref{fig1} (c), most annotations in Charades-STA start at the beginning of videos and last roughly 25\% of the video length. This leads to a strong bias toward the frequently annotated locations. If a model is trained to predict the target at the \textit{head} locations (yellow regions in Figure \ref{fig1} (c)) more frequently than at the \textit{tail} locations (green regions), the former is more likely to prevail over the latter during testing.
\item \textit{High correlation between user queries and moment locations}. In Figure \ref{fig1} (d) and (e), the queries including a verb "open" mostly match target moments at the beginning of videos. While those corresponding to "leave" are often located at the end. That makes user queries and moment locations spuriously correlated.
\end{itemize}
Although the temporal location bias misleads the model prediction severely, it also has some good, \textit{e.g.,} the temporal location of a moment helps to better model the temporal context for answering the query with temporal languages~\cite{hendricks2018localizing}. Therefore, {the key of VMR is how to properly keep the good and remove the bad effect of moment location for a robust cross-modal matching. As far as we know, no similar effort is devoted to address this critical issue.} 

In this paper, we present a causal VMR framework that builds structural causal model~\cite{pearl2016causal} to capture the causality between different components in VMR. The causal graph is depicted in Figure \ref{fig1} (b) which consists of four variables: $Q$ (query), $V$ (video moment), $Y$ (prediction), and $L$ (moment location). The prediction of traditional models is inferred by using the probability $P(Y|Q,V)$. In the causal graph, $L$ is a hidden confounder~\cite{pearl2018book} that spuriously associates ${V}$ and $Y$, which has been long ignored by traditional VMR models. The effect of $L$ on $Y$ mainly comes from the biases in datasets, while the effect of $L$ on $V$ is mainly due to the entanglement of latent location factor in the representation of $V$. Here, the confounder $L$ opens a backdoor path: ${V}$$\leftarrow$$L$$\rightarrow$$Y$ that hinders us from finding the true effect on $Y$ only caused by $(Q, V)$. 
To remove the harmful confounding effects \cite{pearl2018book} caused by moment location, we design a \textbf{Deconfounded Cross-modal Matching} (DCM) method. It first disentangles the moment representation to infer the core feature of visual content, and then applies causal intervention on the multimodal input based on backdoor adjustment \cite{pearl2016causal}. Specifically, we replace $P(Y|Q,V)$ with $P(Y|do(Q, V))$ by \textit{do} calculus which encourages the model to make an unbiased prediction. 
The query is forced to fairly interact with all possible locations of the target based on the intervention. We expect the deconfounding model to be able to robustly localize the target even when the distribution of moment annotations is significantly changed. This triggers another key question: {how to evaluate the generalizability of VMR models?} 

In general, the split of training and testing sets follows the independent and identically distributed (IID) assumption \cite{vapnik1999overview}. Due to the existence of temporal location biases in the datasets, the IID testing sets are insufficient to test the causality between ${(Q, V)}$ and $Y$. To address this issue, we introduce the \textbf{Out-Of-Distribution} (OOD) testing \cite{teney2020value} into evaluation. Our basic idea is to modify the location distribution of moment annotations in the testing sets by inserting a short video at the beginning/end of testing videos. It is only applied on testing sets, which does not introduce any new biases that affect the model training. 
 Extensive experiments on both IID and OOD testing sets of {ActivityNet-Captions}~\cite{krishna2017dense}, {Charades-STA} \cite{gao2017tall}, and {DiDeMo}~\cite{anne2017localizing}, clearly demonstrate that DCM not only achieves higher retrieval performance but also endows the model with strong generalizability against distribution changes. 
 
 In summary, this work makes the following main contributions:
 \begin{itemize}
 \item We build a causal model to capture the causalities in VMR and point out that the moment temporal location is a hidden confounding variable that hinders the pursuit of true causal effect.
 \item We propose a new method, DCM, for improving VMR, that applies causal intervention on the multimodal model input to remove the confounding effects of moment location.
 \item We conduct extensive experiments on the IID and OOD testing sets of three benchmark datasets, which show the ability of DCM to perform accurate and robust moment retrieval.
\end{itemize}

\section{Causal View of VMR}
In this section, we first formally formulate the task of VMR and then introduce a causal graph to clearly reveal how the confounder $L$ spuriously correlate the multimodal input and prediction.
\vspace{-0.5em}
\subsection{Problem Formulation}\label{formulation}
Given a query and a video with temporal length $\tau$, our task is to identify a target moment $\hat{v}$ in the given video, 
starting at timestamp $\tau_s$ and ending at timestamp $\tau_e$ ($\tau_s< \tau_e<\tau$), which semantically matches the query ${q}$. 
The given video is first processed into a set of video moment candidates $\{v_i\}_{i=1}^N$, with diverse temporal durations, either by multi-scale sliding window~\cite{gao2017tall,liu2018attentive,liu2018cross} or hierarchical pooling~\cite{zhang2019learning}, where $N$ is the number of moment candidates. Each moment $v$ is indexed by a temporal location $\ell$, \textit{i.e.}, a pair of timestamps $(t_s, t_e)$. 
 We compute the relevance score ${y}$ (\textit{e.g.}, IoU score) between each moment candidate $v$ and target $\hat{v}$ as the supervision for training, where $y$ can be either binary $\{0,1\}$ or dense $\left(0,1\right)$. The candidates with $y\geq 0.5$ are usually treated as positive. As such, the task of video moment retrieval can be formally defined as:
\begin{itemize}
	\item \textbf{Input}: A corpus of query-moment pairs and their relevance scores: $\{\mathcal{Q}, \mathcal{V}, \mathcal{Y}\}$, where $\mathcal{Q}, \mathcal{V}$, and $\mathcal{Y}$ denote the sets of queries, moments, and relevance scores, respectively.
	\item \textbf{Output}: A cross-modal matching function $f$: $\mathcal{Q}\times\mathcal{V}\rightarrow\mathbb{R}$ that maps each query-moment pair to a real value by effectively modeling intra-modality  and inter-modality feature interactions. During inference, we rank the moment candidates based on the estimated matching scores, and return the best candidate to user.
\end{itemize} 
The solution usually involves: 1) learning the representation of query $\mathbf{q}\in\mathbb{R}^d$ and moment $\mathbf{v}\in\mathbb{R}^d$, and 2) modeling the cross-modal relationship. We mainly focus on the second point. In particular, we will investigate how the temporal locations of video moments spuriously affect the prediction $f(\mathbf{q}, \mathbf{v})$, and then aim to remove the {spurious correlation} caused by the moment location.
\vspace{-0.5em}
\subsection{Causal Graph}\label{CausalGraph}
We describe the causalities among four variables in VMR: query $Q$, video moment $V$, moment location $L$, and model prediction $Y$ with a simple causal graph in Figure \ref{fig1} (b). The direct link denotes the causality between two nodes: cause$\rightarrow$effect. The causal graph is a directed acyclic graph that indicates how the variables $\{Q, V, L, Y\}$ interact with each other through the causal links. 
Traditional VMR models, \textit{e.g.}, \cite{gao2017tall,zhang2019learning}, have only two links: $Q$$\rightarrow$$Y$ and $V$$\rightarrow$$Y$, and predict the target by joint probability $P(Y|Q,V)$.  In our causal graph, the moment location $L$ is a confounder~\cite{pearl2016causal,pearl2018book} that influences both the correlated variable $Y$ and independent variable $V$, leading to spurious correlation between $V$ and $Y$:
\begin{itemize}
	\item $L$$\rightarrow$$Y$ is rooted from the frequency prior $P(Y|L)$. Dataset annotators prefer to select short moments at specific locations of videos as the target~\cite{anne2017localizing}, \textit{e.g.}, the beginning or end of videos, as depicted in Figure \ref{fig1} (c), which leads to a \textit{long-tailed} location distribution of moment annotations. It introduces a harmful causal effect that misleads the prediction of targets at the \textit{tail} (green regions) location biased towards the data-rich \textit{head} (yellow regions) locations.	
	The location $L$ leads to not only the direct effect $L$$\rightarrow$$Y$ but also the joint effect $(Q, L)$$\rightarrow$$Y$, which will be investigated in section \ref{exp}. 	
	\item $L$$\rightarrow$$V$ denotes the effect of $L$ on moment representation. In VMR, the temporal location of moments leads to a temporal context prior that affects the representation learning of moments in videos. {The extracted moment representation from a backbone network is usually {entangled} with a latent location factor.} Although it enriches the representation with location-aware features, it introduces the harmful spurious correlation between $(Q, V)$ and $Y$ via the backdoor path: $V$$\leftarrow$$L$$\rightarrow$$Y$. Similar feature entanglements are common in video domain~\cite{denton2017unsupervised}. 
\end{itemize}
So far, we have clearly revealed how the moment location $L$ confound $V$ and $Y$ via the backdoor path that makes the effect of visual content largely overlooked, as reported by \cite{otani2020uncovering}. Next, we propose a deconfounding method to remove the confounding effects.
\vspace{-0.4em}
\section{METHODOLOGY}\label{methodology} 

\begin{figure}[tb]
	\centering
	\includegraphics[width=3.2in]{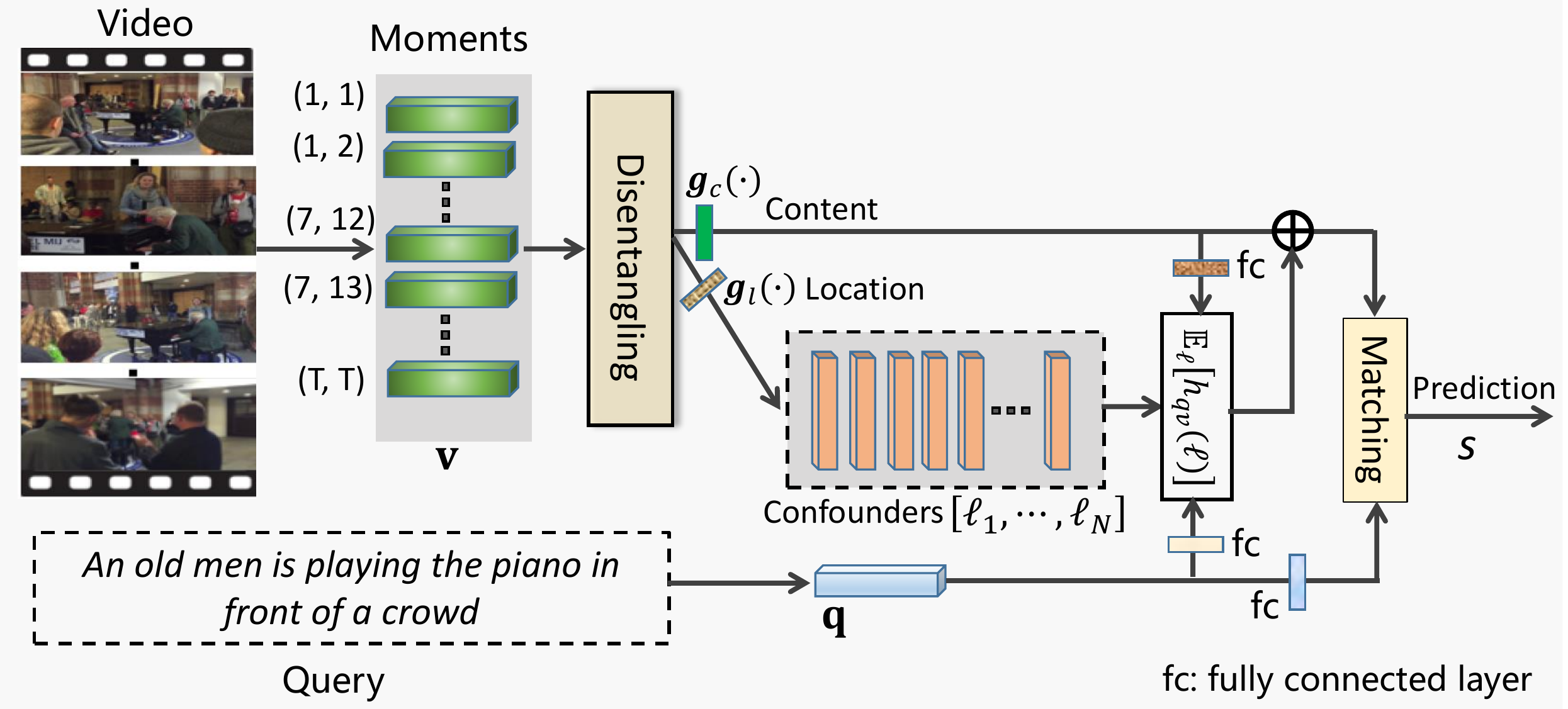}
	\vspace{-0.15in}
	\caption{A brief overview of our deconfounded cross-modal matching for the task of VMR.}
	\vspace{-0.2in}
	\label{fig2}
\end{figure}

Given the representation of query $\mathbf{q}$ and a moment candidate $\mathbf{v}$, our goal is to predict the matching score $s$ between the candidate and the query by a deconfounding method, DCM. As depicted in Figure \ref{fig2}, it is composed of two key steps to achieve the deconfounding: 1) feature disentangling on ${V}$ that infers the core feature to represent the visual content of moment candidate, reducing the effect of $L$ on $\mathbf{v}$; 2) causal intervention on the multimodal input to remove the harmful confounding effects of $L$ from $Y$.
\vspace{-0.4em}
\subsection{Moment Representation Disentangling}\label{disentangled}
As discussed before, the moment representation is usually entangled with a location factor. To achieve the deconfounding, the first step is to ensure that the representations of $V$ and $L$ are independent with each other. In this section, we propose to disentangle $\mathbf{v}$ into two independent latent vectors: content $\mathbf{c}^v\in\mathbb{R}^d$ and location $\bm{\ell}^v\in\mathbb{R}^d$: 
\begin{equation}
\mathbf{c}^v = g_c(\mathbf{v}), \bm{\ell}^v = g_{\ell}(\mathbf{v}), 
\end{equation} 
where $g_c(\cdot)$ and $g_\ell(\cdot)$ can be implemented by two fully connected layers. To achieve the disentangling, we introduce two learning constraints to optimize the parameters of $g_c(\cdot)$ and $g_\ell(\cdot)$.
\begin{itemize}
	\item \textbf{Reconstruction}. Since we can access the original location $(t_s, t_e)$ of each moment $v$ during training, we use it to supervise $g_\ell(\cdot)$ by a reconstruction loss $\mathcal{L}_{recon}(\bm{\ell}^v, \mathbf{p})$,  where $\mathbf{p}\in\mathbb{R}^d$ is a non-learnable positional embedding vector~\cite{vaswani2017attention} of $(t_s, t_e)$. $\mathcal{L}_{recon}(\cdot,\cdot)$ can be any reconstruction losses that force $\bm{\ell}^v$ to approximate the real location feature. 
	Different from existing efforts \cite{anne2017localizing,mun2020LGI,zeng2020dense} that exploit $(t_s, t_e)$ as the side information to augment the moment representation, we focus on disentangling $\mathbf{v}$ to capture the core effect of visual content and remove the confounding effects of moment location. Injecting extra location feature into $\mathbf{v}$ will obviously exacerbate the correlation between $L$ and $V$.
	\item \textbf{Independence Modeling}. Different from $\bm{\ell}^v$, there is no supervision on the content $\mathbf{c}^v$. Besides, $\mathbf{c}^v$ and $\bm{\ell}^v$ are both estimated from $\mathbf{v}$, they do not immediately satisfy the \textit{independence} requirement. We introduce an independence learning regularizer  $\mathcal{L}_{indep}(\bm{\ell}^v, \mathbf{c}^v)$ that forces $\bm{\ell}^v$ to be independent with $\mathbf{c}^v$ in the latent space. $\mathcal{L}_{indep}(\cdot,\cdot)$ can be any statistical dependence measures, \textit{e.g.}, distance correlation and mutual information. 
\end{itemize}
We will describe the details of $\mathcal{L}_{indep}$ and $\mathcal{L}_{recon}$ in section \ref{losses}. The representation disentangling on $V$ reduces the correlation between $V$ and $L$, which cuts off the link $L$$\rightarrow$$V$ in the feature level. The next question is how to remove the harmful confounding effects of $L$ from $Y$ for making a robust prediction on each candidate.

\vspace{-0.5em}
\subsection{Causal Intervention}\label{intervention}
\begin{figure}[tb]
	\centering
	\includegraphics[width=2.2in]{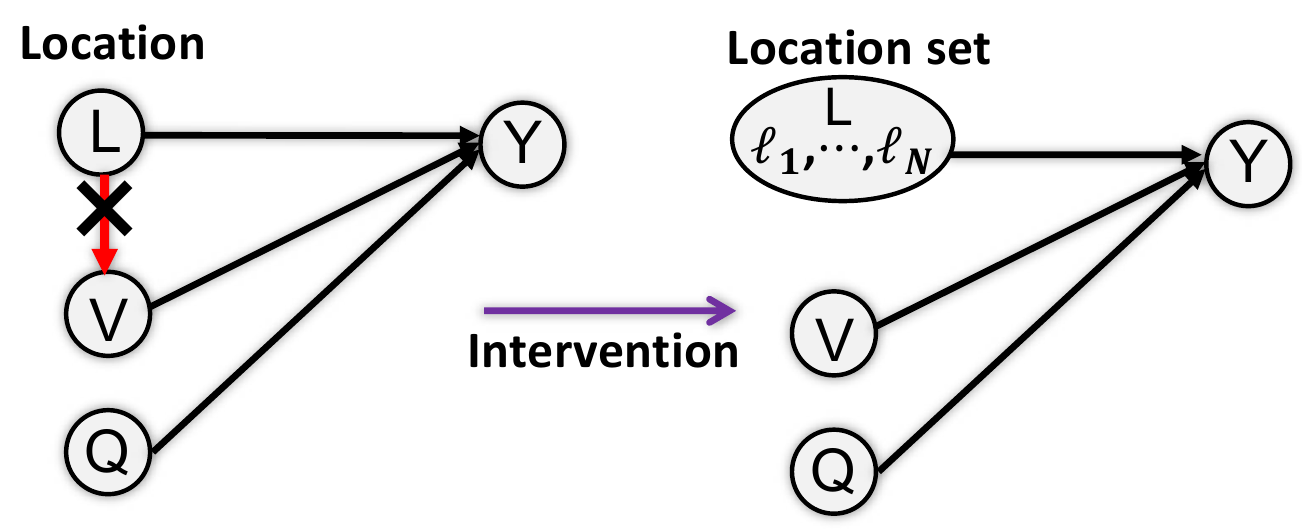}
	\vspace{-0.15in}
	\caption{A brief illustration of our causal intervention.}
	\vspace{-0.2in}
	\label{fig3}
\end{figure}

As described before, $L$ is a confounder that leads to spurious correlation between $V$ and $Y$. For example, in an unbalanced VMR dataset, the high-frequency location $\ell_h$, has higher chance to be selected as the target location than the low-frequency location $\ell_l$: $P(Y|\ell_h)$$>$$P(Y|\ell_l)$. That will mislead the model towards predicting a higher score on the candidate $v_{\ell_h}$ at $\ell_h$ than the candidate $v_{\ell_l}$ at $\ell_l$, without truly looking into the details of multimodal input. Even if the query is considered, if not removing the confounding effects of moment location, the model will still be easily misled by the query-specific location prior (Figure \ref{fig1} (d) and (e)). 
In this section, {{as shown in Figure \ref{fig3}}}, we propose to use $do$-calculus~\cite{pearl2018book} to intervene the model input based on back-door adjustment. 
By applying the Bayes rule on the new casual graph, we have 
\begin{equation}\label{do-operation}
P\left(Y|do\left(Q, V\right)\right) = P\left(Y|Q, do\left(V\right)\right) = \sum_{{\ell}\in L} P\left(Y|Q, V, \ell\right) P({\ell}),
\end{equation}
where $P({\ell})$ is the prior of location. We see from Eq. (\ref{do-operation}) that the aim of the \textit{intervention} is to force the query to fairly interact with all the possible locations of the target moment for making an unbiased prediction, subject to the prior $P({\ell})$. In this way, the model is prevented from memorizing the corresponding locations of target moments during training. Note that we force the prior $P({\ell})$ to be a constant $\frac{1}{N}$, following the assumption that each location has equal opportunity to be the target location. For the location confounder set $L$, since we can hardly enumerate all the moment locations in the real world, we approximate $L$ as the location set of all sampled moment candidates, \textit{i.e.}, $L=\{{\ell}_k\}_{i=1}^N$, where ${{\ell}_k}$ is represented by ${\bm{\ell}_k}\in\mathbb{R}^d$, \textit{i.e.}, the disentangled location feature from $\mathbf{v}_k$. 

Given the representations of the query and a moment candidate, Eq. (\ref{do-operation}) is implemented as $\sum_{\bm{\ell}}p(y|\mathbf{q}, \mathbf{v}, \bm{\ell})P(\bm{\ell})$,~\footnote{To keep section \ref{intervention} self-contained, we still use $\mathbf{v}$ to denote the moment representation for simplicity. It can either be the original or disentangled representation of $V$.} where $p(y|\mathbf{q}, \mathbf{v}, \bm{\ell})$ is the output of a cross-modal matching network $f(\cdot)$:
\begin{equation}\label{cross-modal-matching}
p(y|\mathbf{q}, \mathbf{v}, \bm{\ell})=\sigma(f(\mathbf{q}, \mathbf{v}, \bm{\ell})),
\end{equation}
where $\sigma(\cdot)$ is the sigmoid function $\sigma(x)=\frac{1}{1+\textrm{exp}(-x)}$ that transforms the output of $f(\cdot)$ into $(0, 1)$. In summary, the implementation of Eq. (\ref{do-operation}) is formally defined as
\begin{equation}\label{expectationE_l}
P\left(Y|do\left(Q, V\right)\right) :=\mathbb{E}_{\bm{\ell}}\left[\sigma(f(\mathbf{q}, \mathbf{v}, \bm{\ell}))\right].
\end{equation}
To calculate Eq. (\ref{expectationE_l}), we found that $\mathbb{E}_{\bm{\ell}}$ needs expensive sampling. Fortunately, following recent efforts~\cite{yue2020interventional,zhang2020causal}, we can first adopt the Normalized Weighted Geometric Mean (NWGM)~\cite{xu2015show} to approximately move outer expectation $\mathbb{E}_{\bm{\ell}}$ into the sigmoid function:
\begin{equation}\label{sigma(E_l)}
P\left(Y|do\left(Q, V\right)\right) :=\mathbb{E}_{\bm{\ell}}\left[\sigma(f(\mathbf{q}, \mathbf{v}, \bm{\ell}))\right] \approx \sigma\big(\mathbb{E}_{\bm{\ell}}\left[f(\mathbf{q}, \mathbf{v}, \bm{\ell})\right]\big).
\end{equation}
The further calculating of Eq. (\ref{sigma(E_l)}) depends on the implementation of the matching network $f(\cdot)$. 
\textit{If $f(\cdot)$ is a linear model, we can further move the expectation into $f(\cdot)$, \textit{i.e.}, $\mathbb{E}(f(x))=f(\mathbb{E}(x))$}. Then we just need one forward pass to obtain the prediction. Besides, to seek the true effect of visual content of $V$ on $Y$, the representation $\mathbf{v}$ is implemented as the disentangled moment representation. We will describe the detailed implementation of $f(\cdot)$ in section \ref{dcm_imp}. So far, we have finished the causal intervention via backdoor adjustment~\cite{pearl2016causal,pearl2018book}, thus closing the backdoor path $V \leftarrow L\rightarrow Y$.

\noindent{\textbf{Summary}}:  From Eq. (\ref{sigma(E_l)}), we can see that the effect on $Y$ comes from $Q, V$, and all the possible locations of the target moment, \textit{i.e.}, $\{{\ell}_k\}_{i=1}^N$, which prevents the model from exploiting the location priors $P(Y|L)$ and $P(Y|Q, L)$. The disentangling module described in section \ref{disentangled} supports the deconfounding in the \textit{feature} level while the intervention module contributes more in the \textit{prediction} level. Both of them are indispensable for achieving the deconfounding. 
\vspace{-0.6em}
\subsection{Model Implementation}\label{Implementation}
This section presents the model implementation. It mainly consists of the moment candidate sampling and encoding, query encoding, cross-modal matching networks $f(\cdot)$, and the learning objective.
\vspace{-0.5em}
\subsubsection{\textbf{Moment Sampling and Encoding}} For a given video, we first segment it into a sequence of $T$ non-overlapping video clips by fixed-interval sampling. 
The clip feature can be obtained by applying pretrained CNNs, such as C3D~\cite{tran2015learning} and I3D~\cite{carreira2017quo}, on video frames. Then we enumerate all moment candidates from the given video by composing any ordered lists of video clips. Each clip can be also treated as a short moment. The representation $\mathbf{v}\in \mathbb{R}^d$ of each candidate is obtained by applying a stacked pooling network over the set of video clips, following  \cite{zhang2019learning}. All the feature vectors of sampled moment candidates can be encoded into a moment feature tensor $\mathbf{V}\in \mathbb{R}^{d\times T \times T}$. The last two dimensions of $\mathbf{V}$ index the start and end coordinates of all sampled candidates.

\subsubsection{\textbf{Query Encoding}} For the encoding of the query sentence $q$, we first extract the word embeddings by the pretrained Glove word2vec model~\cite{jeffreypennington2014glove}. We then use a three-layer unidirectional LSTM to capture the sequential information in $q$. The query representation $\mathbf{q}\in \mathbb{R}^d$ is the last hidden state of the LSTM output.

\subsubsection{\textbf{Deconfounded Cross-modal Matching (DCM)}} \label{dcm_imp}
Our proposed DCM method can be implemented with most existing matching architectures in VMR (proposals-based). In this work, we implement it with two kinds of cross-modal matching networks $f(\cdot)$:
\begin{itemize}
	\item {Context-aware Multi-modal Interaction} (CMI). It is modified from the fusion module in \cite{gao2017tall}. We implement $f(\cdot)$ with CMI as: 
	\begin{equation}\label{CMI}
	f(\mathbf{q}, \!\mathbf{v}, \!\bm{\ell})\!=\!\mathbf{w}^T\Big(\mathbf{W}_1\Big(\big(\bar{\mathbf{q}} \!\odot \!\phi_{\mathbf{V}}\big(\bar{\mathbf{v}}\!+\!{h}_{qv}(\bm{\ell})\big)\!\big) \!\oplus\! \big(\bar{\mathbf{q}} \!+ \!\phi_{\mathbf{V}}\big(\bar{\mathbf{v}}\!+\!{h}_{qv}(\bm{\ell})\big)\big)\Big)\!\Big),
	\end{equation}
	where $\bar{\mathbf{v}}$ and $\bar{\mathbf{q}}$ denote the transformed moment and query representations, respectively: $\bar{\mathbf{v}}=\mathbf{c}^v + \mathbf{W}_2 \bm{\ell}^v$ and  $\bar{\mathbf{q}}=\mathbf{W}_3 \mathbf{q}$. The $\mathbf{w}\in\mathbb{R}^{d}$,  $\mathbf{W}_1\in\mathbb{R}^{d\times 2d}$, $\mathbf{W}_2\in\mathbb{R}^{d\times d}$, and $\mathbf{W}_3\!\in\!\mathbb{R}^{d\times d}$ denote the learnable parameters of four fully-connected layers, respectively. The $\odot$ denotes the element-wise multiplication and $\oplus$ denotes the channel-wise vector concatenation.  ${h}_{qv}(\bm{\ell})$ is a feature transformation of $\bm{\ell}$, parameterized by the features of $q$ and $v$. 
	 $\phi_{\mathbf{V}}(\bar{\mathbf{v}}+{h}_{qv}(\bm{\ell}))$ denotes a convolutional operation on the intervened moment tensor $\mathbf{V}'\!\in\! \mathbb{R}^{d\times T \times T}$, where each element has been changed from $\mathbf{v}$ to $\bar{\mathbf{v}}+{h}_{qv}(\bm{\ell})$, to aggregate the pre-context and post-context moment features~\cite{gao2017tall} into current moment $v$, with the weight parameter $\mathbf{W}_4 \in\mathbb{R}^{d\times d\times K \times K}$ ($K$ denotes the kernel size). We remove the non-linear activation function behind the convolution operation. 
	 Then, we {replace ${h}_{qv}(\bm{\ell})$ in Eq. (\ref{CMI}) with $\mathbb{E}_{\bm{\ell}}\big[{h}_{qv}\big(\bm{\ell}\big)\big]$} to approximately compute $\mathbb{E}_{\bm{\ell}}\big[f(\mathbf{q}, \mathbf{v}, \bm{\ell})\big]$.
	\item {Temporal Convolutional Network} (TCN)~\cite{zhang2019learning}. It is a SOTA matching network in VMR. We implement TCN-based $f(\cdot)$ as:
	\begin{equation}\label{TCN}
	f(\mathbf{q}, \mathbf{v}, \bm{\ell})=\mathbf{w}^T\Big(\phi^*_{\mathbf{V}}\Big(\bar{\mathbf{q}} \odot \big(\bar{\mathbf{v}}+{h}_{qv}(\bm{\ell})\big)\Big) \Big),
	\end{equation}
	where $\phi^*_{\mathbf{V}}(\cdot)$ denotes a multi-layer 2D CNN over the intervened moment tensor $\mathbf{V}'\in \mathbb{R}^{d\times T \times T}$ that captures the temporal dependencies between adjacent moments into current moment $v$. Eq. (\ref{TCN}) is non-linear due to the rectified linear unit $\textit{Relu} (\cdot)$  behind each convolutional layer. Fortunately, based on the theoretical analysis of \cite{baldi2014dropout} on the rectified linear unit, the approximation $\mathbb{E}(\textit{Relu} (x))\approx \textit{Relu}\left(\mathbb{E}\left[x\right]\right)$ still holds. Then we can obtain 
	\begin{equation}\label{TCN2}
	\mathbb{E}_{\bm{\ell}}\left[f(\mathbf{q}, \mathbf{v}, \bm{\ell})\right] \approx \mathbf{w}^T\Big(\phi^*_{\mathbf{V}}\Big(\bar{\mathbf{q}} \odot \big({\bar{\mathbf{v}}+\mathbb{E}_{\bm{\ell}}\left[{h}_{qv}\big(\bm{\ell}\big)\right]}\big)\Big) \Big).
	\end{equation}
\end{itemize}
The key of the two above-mentioned cross-modal matching networks $f(\cdot)$ is to compute $\mathbb{E}_{\bm{\ell}}\left[{h}_{qv}\big(\bm{\ell}\big)\right]$. We implement ${h}_{qv}\big(\bm{\ell}\big)$ as the scaled Dot-Product attention \cite{vaswani2017attention} to adaptively assign weights on different location confounders in $\mathbf{L}=\left[\bm{\ell}_1, \cdots, \bm{\ell}_N\right]\in \mathbb{R}^{N\times d}$ with specific input of query $q$ and moment $v$. Then we have
\begin{equation}\label{attention}
\mathbb{E}_{\bm{\ell}}\left[{h}_{qv}\big(\bm{\ell}\big)\right]=\sum_{\bm{\ell}}\big[\mathrm{softmax}\big(\mathbf{K}^T\mathbf{m}/\sqrt{d}\big)\odot {\mathbf{L}'}\big]P(\bm{\ell}),
\end{equation}  
where $\odot$ denotes the element-wise product that supports broadcast, and $\mathbf{m}=\mathbf{W}_5\mathbf{q} + \mathbf{W}_6\mathbf{c}^v$, $\mathbf{K}=\mathbf{W}_7\mathbf{L}^T$, and $\mathbf{L}'=\mathbf{L}\mathbf{W}_2^T$ with learnable parameters $ \mathbf{W}_5, \mathbf{W}_6, \mathbf{W}_7$, and $\mathbf{W}_2\in\mathbb{R}^{d\times d}$. 

So far, we have introduced the implementation of the proposed DCM. Note that, we do not use the feature of original moment location $(t_s, t_e)$ in the inference stage. 
Given the representation of query $\mathbf{q}$ and moment candidate $\mathbf{v}$, we can predict the deconfounded cross-modal matching score as $s=\sigma\big(\mathbb{E}_{\bm{\ell}}\left[f(\mathbf{q}, \mathbf{v}, \bm{\ell})\right]\big)$, where the confounder vector $\bm{\ell}$ is disentangled from moment representation.
\subsubsection{\textbf{Learning Objective}} \label{losses}To train the DCM-based VMR methods, we use the scaled Intersection over Union (IoU) score  \cite{zhang2019learning} between the locations of moment candidate $v$ and groundtruth $v_{gt}$ as the supervision $y$. The basic loss is the binary cross entropy loss:
\begin{equation}\label{BCEloss}
\mathcal{L}_{bce}(\mathbf{q},\mathbf{v}, y) = -\big(y\log(s) + (1-y)\log(1-s)\big).
\end{equation}
In this work, we not only train our method with positive query-video training pairs but also exploit the negative query-video pairs as \textbf{counterfactual} training data to optimize the network parameters. Existing efforts all assume that the existence of target moment in the given video is true, forgoing teaching the model the ability of \textit{counterfactual thinking}, \textit{i.e., What if the target moment does not exist in the given video?}. By introducing the \textit{counterfactual thinking}, we expect the learned model to focus more on the content of video and query, rather than the location of moment candidates. 
In summary, our model is trained with a linear combination of four losses:
\begin{equation}\label{full_loss}
\mathcal{L}=\mathcal{L}_{bce}^+ + \mathcal{L}_{bce}^- + \lambda_1 \mathcal{L}_{recon} + \lambda_2 \mathcal{L}_{indep},
\end{equation}
where the \textbf{counterfactual loss} $\mathcal{L}_{bce}^-$ is computed over negative query-video pairs with the supervision $y=0$. $\mathcal{L}_{recon}$ and $\mathcal{L}_{indep}$ are two learning terms for feature disentangling in section \ref{disentangled} with two hyperparameters $\lambda_1$ and $\lambda_2$. We implement the reconstruction term by $\mathcal{L}_{recon}=\lVert \bm{\ell}^v - \mathbf{p} \rVert_2$ for its simplicity. The independence term is implemented based on distance correlation~\cite{szekely2007measuring}: $\mathcal{L}_{indep}={\textrm{dCov}(\bm{\ell}^v, \mathbf{c}^v)}/{\sqrt{\textrm{dVar}(\bm{\ell}^v)   \cdot \textrm{dVar}(\mathbf{c}^v) }}$, 
where $\textrm{dCov}(\cdot)$ and $\textrm{dVar}(\cdot)$ denote the distance covariance and variance, respectively.
The overall loss is computed as the average over a training batch.
\section{Experiments}\label{exp}
 In this section, we evaluate the effectiveness of our DCM methods by extensive comparison with baseline methods (CMI and TCN) and SOTA VMR methods in both IID and OOD settings. 
 \vspace{-0.5em}
\subsection{Datasets and Experimental Setting}
\subsubsection{\textbf{Datasets}}
 We validate the performance of DCM methods on three public datasets: \textbf{ActivityNet-Captions} (ANet-Cap)~\cite{krishna2017dense}, \textbf{Charades-STA}~\cite{gao2017tall}, and \textbf{DiDeMo}~\cite{anne2017localizing}: 
\noindent{(1) \textit{ANet-Cap}} depicts diverse human activities in daily life, consisting of about 20k videos taken from the ActivityNet caption dataset~\cite{krishna2017dense}. We follow the setup in \cite{zhang2019cross,zhang2019learning,NIPS2019_8344} to split the dataset;
\noindent{(2) \textit{Charades-STA}} is constructed for VMR by \cite{gao2017tall} based on the Charades video dataset~\cite{sigurdsson2016hollywood}. The videos are about daily indoor activities. Currently, it is the most popular VMR dataset;
\noindent{(3) \textit{DiDeMo}} is prepared by \cite{anne2017localizing} with more than 10k videos from YFCC100M. It depicts diverse human activities in real world. The detailed dataset statistics are summarized in Table \ref{dataset_statistics}.

\begin{table}[tb]
	\renewcommand{\arraystretch}{0.97}
	\renewcommand\tabcolsep{3.4pt}
	\caption{The statistics of three public VMR datasets.}
		\vspace{-0.15in}
		\centering
					\scalebox{0.87}{%
		\begin{threeparttable}
		\begin{tabular}{c|c|c|c|c}
			\hline
			\hline
			Dataset & $\#$Videos &$\#$Anno. (train/val/test) &$\overline{L}_{\textrm{vid}}$ &$\overline{L}_{\textrm{mom}}\pm{\triangle}_{\textrm{mom}}$\\
			\hline
			ANet-Cap&19,207 &37,417/17,505/17,031 &117.6s&36.2$\pm40.2$s\\
			Charades-STA&6,672 & 12,408/-/3,720&30.6s&8.2$\pm 3.6$s\\
			DiDeMo&10,464 &132,233/16,765/16,118  &30s&6.85$\pm3.5$s\\
			\hline			
		\end{tabular}
    \begin{tablenotes}
	\footnotesize
	\item[1] $\#$Anno. denotes the number of query-moment annotation pairs in different sets (train/val/test). 
	\item[2] $\overline{L}_{\textrm{vid}}$  and $\overline{L}_{\textrm{mom}}$ denote the average lengths of videos and moments, respectively. 
	\item[3] ${\triangle}_{\textrm{mom}}$ denotes the standard deviation of moment length.
	\item[4] In DiDeMo~\cite{anne2017localizing}, each query may have multiple moment annotations.
\end{tablenotes}
\end{threeparttable}}
\vspace{-0.12in}
\label{dataset_statistics}
\end{table}

\begin{table*}[tb]
	\renewcommand{\arraystretch}{0.9}
	\renewcommand\tabcolsep{2.8pt}
	\definecolor{white}{rgb}{1,1,1}
	\caption{Performance comparison (R@1,\%) with state-of-the-arts (SOTAs) and baselines (Bases) on three datasets (The higher the better). The $\dagger$  and $\ddagger$ denote relative improvements larger than $5\%$ and $10\%$ \textit{w.r.t.} R@1, respectively. The $\ast$ and $\star$ denote the statistical significance \textit{w.r.t.} mIoU for $p<0.05$ and $p<0.01$, respectively, compared with the baseline counterparts.}
	\vspace{-0.12in}
	\centering
	\scalebox{0.86}{%
		\begin{threeparttable}		
			\begin{tabular}{c|{c}||*{2}{l}{l}||*{2}{l}{l}||*{2}{l}{l}||*{2}{l}{l}||*{2}{l}{l} } 
				\hline
				&\multirow{3}{*}{\textbf{Method}} &
				\multicolumn{3}{c||}{\textbf{ANet-Cap} (C3D)}&\multicolumn{3}{c||}{\textbf{Charades-STA}(C3D)}&\multicolumn{3}{c||}{\textbf{Charades-STA}(I3D)}&\multicolumn{3}{c||}{\textbf{Charades-STA}(VGG)}&\multicolumn{3}{c}{\textbf{DiDeMo} (Flow)}\\
				\cline{3-16} 
				&	&\multicolumn{2}{c}{\textbf{IoU>m}} &
				\multirow{2}{*}{\textbf{mIoU}} 
				&\multicolumn{2}{c}{\textbf{IoU>m}} &
				\multirow{2}{*}{\textbf{mIoU}} 
				&\multicolumn{2}{c}{\textbf{IoU>m}} &
				\multirow{2}{*}{\textbf{mIoU}} 			
				&\multicolumn{2}{c}{\textbf{IoU>m}} &
				\multirow{2}{*}{\textbf{mIoU}}            
				&\multicolumn{2}{c}{{\textbf{IoU>m}}} &
				\multirow{2}{*}{\textbf{mIoU}}\\			
				\cline{3-4} \cline{6-7}		
				\cline{9-10}  \cline{12-13}  \cline{15-16} 
				&& 0.5   & 0.7 &  & 0.5   & 0.7  &  & 0.5   & 0.7 &  & 0.5   & 0.7   && 0.7   & 1.0 &  \\ 
				\hline
				\hline
				\multirow{8}{*}{\rotatebox{90}{\textbf{SOTAs (IID)}} }	
				&    MAN\cite{zhang2019man}                     & -& -&-        &-   &-        &-          & - & - &  -       &46.5 &22.7 &-       & - &27.0 &41.2  \\  
				&  CBP \cite{wang2019temporally}             & 35.8 & 17.8  & 36.9       &36.8 &18.9   &35.7   & - & - &              & - & - &  -          & - &- &  -  \\  
				&	SCDM \cite{NIPS2019_8344}                & 36.8  & 19.9   &  -               &-       & -       &-          &54.4&33.4 & -   & - & - &   -         & - &- &  -  \\  
				&    2D-TAN\cite{zhang2019learning}        & 44.5& {26.5} &43.2            &47.0   &27.2  &42.0          & 53.7&31.2 & 47.0            &42.8 &23.3 &- &35.3 &25.6 &47.8  \\  
				& 	DRN \cite{zeng2020dense}                  &\textbf{45.5} &24.4 &-   &45.4 &26.4  &-           & 53.1 & 31.8 & -  &42.9 &23.7 &-  & - &- & -   \\  
				&	LGI \cite{mun2020LGI}                        &43.0 &25.1     &42.6         &50.5  &27.7      &44.9            &59.5 &35.5 &51.4&44.7&24.0 &40.8       & - &- &  -  \\  
				&  VSLNeT \cite{zhang-etal-2020-span}  &43.2 &26.2      &{43.2}       &47.3 &30.2&45.2       & 53.3&33.7 &49.9   & 39.2 &20.8 &40.3        & - &- &  -  \\  
				\hline
				\rowcolor{gray!12}
				\hline
				\hline
				\cellcolor{white}&	Freq-Prior                 &29.7 &13.9    &32.0        & 29.7    &16.3 &30.1        & 29.7 &16.3  &30.1      & 29.7 &16.3  &30.1        &23.3&19.4 &31.9  \\  
				\rowcolor{gray!12}
				\cellcolor{white}&		Blind-TAN          &45.3 &\textbf{28.6}    &\textbf{43.9}$\pm 0.4$         & 40.2    &23.5   &36.7$\pm 0.2$  & 40.2    &23.5   &36.7$\pm 0.2$   & 40.2    &23.5   &36.7$\pm 0.2$          &24.4 &19.4 &34.5$\pm 1.4$  \\  
				\cline{2-17} 
				&		CMI           & 40.6 &23.2&41.1$\pm0.3$            &43.3&24.6&40.2$\pm0.4$     &48.5&28.1&43.6$\pm0.5$      &40.1&21.9&37.7$\pm 0.4$    &32.9 &25.1 &44.1$\pm 0.5$    \\  
				&		\textbf{CMI+DCM}       &41.4&23.9 &41.6$\pm0.3^\ast$          &52.8$^\ddagger$ & 32.1$^\ddagger$  &47.0$\pm0.4$$^{\star}$       &57.5$^\ddagger$&34.9$^\ddagger$ &50.4$\pm0.3^\star$       &45.2$^\ddagger$ &26.0$^\ddagger$ & 41.5$\pm 0.4^\star$  &\textbf{37.6}$^\ddagger$&\textbf{27.8}$^\ddagger$ &49.4$\pm 0.5^\star$     \\
				\cline{2-17} 
				&		TCN           &43.3 &26.1    &42.3$\pm0.5$      &48.0    &28.9 &43.1$\pm0.5$          &52.6   &32.3   &46.3$\pm0.4$                &43.1 &25.0 &39.3$\pm 0.8$            &35.1 &26.0 &47.4$\pm 0.7$\\  
				\multirow{-6}{*}{\rotatebox{90}{\textbf{Bases (IID)}}} &\textbf{TCN+DCM}     &44.9 &{27.7}$^\dagger$    &{43.3}$\pm0.2$$^{\star}$  &\textbf{55.8}$^\ddagger$    &\textbf{34.4}$^\ddagger$     &\textbf{48.7}$\pm 0.5$$^{\star}$   &\textbf{59.7}$^\ddagger$& \textbf{37.8}$^\ddagger$&\textbf{51.5}$\pm 0.4^\star$   &\textbf{47.8}$^\ddagger$ &\textbf{28.0}$^\ddagger$&\textbf{43.1}$\pm 0.4^\star$   &37.5$^\dagger$ &27.6$^\dagger$ &\textbf{49.9}$\pm 0.5^\star$  \\  
				\hline
				\hline
				\rowcolor{gray!12}\cellcolor{white}	&	Freq-Prior   &3.9  & 0 &20.6       &0.1 &0    &7.7            &0.1 &0    &7.7           &0.1 &0    &7.7                     &0 & 0 & 0 \\  
				\rowcolor{gray!12}\cellcolor{white}&		Blind-TAN            &16.2 &6.4    &22.0$\pm0.2$       & 15.0   &6.1    &19.2$\pm 0.4$     & 15.0   &6.1    &19.2$\pm 0.4$    & 15.0   &6.1    &19.2$\pm 0.4$     &3.9 &2.1 &6.8$\pm 2.8$  \\
				&		{LGI} \cite{mun2020LGI}        &16.3&6.2 &22.2           & 26.2&11.4 &26.8     &42.1 &18.6 &41.2      &24.1 &8.2 &27.8    & - & - &  -    \\
				&		{VSLNeT} \cite{zhang-etal-2020-span} & - & - &   -          &21.9&12.0 &30.0     &17.5 &8.8 &26.5      & - & - &-    & - & - &  -    \\
				\cline{2-17} 
				&		CMI        &12.3 &5.2    &19.1$\pm 0.5$     &32.1 &15.3 &36.1$\pm 2.5$     &30.4    &16.4  &30.3$\pm 4.2$         &20.5 &8.0 &25.6$\pm 7.4$   &21.4 &18.1 &32.7$\pm0.6$   \\  
				&		\textbf{CMI+DCM}       &16.8$^{\ddagger}$  &6.4$^{\ddagger}$  &23.9$\pm0.3^\star$           & 33.7$^\dagger$ & \textbf{17.2}$^\ddagger$  & 37.4$\pm0.6$     &37.8$^\ddagger$ &\textbf{19.7}$^\ddagger$ &39.6$\pm0.8^\star$       &24.5$^\ddagger$&11.5$^\ddagger$ &31.3$\pm 0.8^\ast$    &29.9$^\ddagger$&23.7$^\ddagger$ &40.0$\pm 2.0^\star$     \\
				\cline{2-17} 
				&		TCN       &16.4  & 6.6 &23.2$\pm0.9$     &30.6  &13.2   &30.8$\pm2.3$      & 27.1    & 13.1 &25.7$\pm3.2$   & 21.8&9.1 &25.9$\pm 1.2$     &29.9 &21.9 &\textbf{42.1}$\pm1.6$   \\  
				\multirow{-8}{*}{\rotatebox{90}{\textbf{OOD-1}}}&		\textbf{TCN+DCM}    &\textbf{18.2}$^{\ddagger}$  & \textbf{7.9}$^{\ddagger}$ & \textbf{24.4}$\pm0.5$$^{\star}$      &\textbf{39.7}$^\ddagger$   &{16.3}$^\ddagger$    & \textbf{39.6}$\pm 0.9$$^{\star}$    &\textbf{44.4}$^\ddagger$&\textbf{19.7}$^\ddagger$& \textbf{42.3}$\pm 1.1^\star$   &\textbf{31.6}$^\ddagger$  &\textbf{12.2}$^\ddagger$&\textbf{34.3}$\pm 0.9^\star$ &\textbf{31.6}$^\dagger$ &\textbf{24.4}$^\ddagger$ &42.0$\pm 1.6$ \\  
				\hline
				\hline
				\rowcolor{gray!12}	\cellcolor{white}&	Freq-Prior      &0.5  &0  & 18.4            &0    &0  &2.5          &0    &0  &2.5         &0    &0  &2.5             & 0& 0&0  \\  
				\rowcolor{gray!12}\cellcolor{white}&		Blind-TAN    &11.1 &3.7    &17.8$\pm0.4$    &13.4    &4.4   &16.1$\pm 0.5$  &13.4    &4.4   &16.1$\pm 0.5$    &13.4    &4.4   &16.1$\pm 0.5$ &3.9 &2.1 &6.7$\pm 2.5$  \\  
				&		{LGI} \cite{mun2020LGI}        &11.0&3.9 &17.3           &20.3 &7.4& 22.1    &35.8&13.5 &37.1      &18.8&5.3&24.3    & - & - & -\\
				&		{VSLNeT} \cite{zhang-etal-2020-span}      & - & - &  -           & 17.4&9.2 &20.7     &10.2 &4.7 &18.4      & - & - &  -    & - & - & -  \\
				\cline{2-17} 
				&		CMI      &10.0& 4.2 &16.8 $\pm0.6$           & 28.5&12.3&35.3$\pm2.4$  &28.1&13.6 &29.0$\pm4.8$       &16.0&5.31 &24.5$\pm 6.9$    &18.3 &15.7 &25.9$\pm 3.4$    \\
				&		\textbf{CMI+DCM}     &\textbf{13.1}&\textbf{5.2} &\textbf{21.6}$\pm0.4$$^{\star}$            & 30.5$^\dagger$ & \textbf{15.2}$^\ddagger$  &35.4$\pm0.8$      &33.2$^\ddagger$ &\textbf{17.1}$^\ddagger$ &36.7$\pm1.1^\star$       &21.1$^\ddagger$ &\textbf{9.6}$^\ddagger$ &29.1$\pm 0.5^\ast$    &29.5$^\ddagger$&23.2$^\ddagger$ &39.3$\pm 2.6^\star$     \\
				\cline{2-17} 
				&		TCN        &11.5  & 3.9 &19.4$ \pm0.9$           & 25.7  & 9.3   &28.4$\pm2.5$      & 21.1    &8.8 &22.5$\pm2.9$           &17.6 &6.2 &22.2$\pm 1.5$    &28.0 &21.2 &\textbf{40.1}$\pm 1.6$    \\
				\multirow{-8}{*}{\rotatebox{90}{\textbf{OOD-2}}}&		\textbf{TCN+DCM}   &  12.9$^{\ddagger}$ &4.8$^{\ddagger}$  & 20.7$\pm0.5$$^\ast$  &\textbf{33.8}$^\ddagger$    &12.4$^\ddagger$    & \textbf{36.3}$\pm 1.1$$^{\star}$    &\textbf{38.5}$^\ddagger$& 15.4$^\ddagger$&\textbf{39.0}$\pm 1.2^\star$   &\textbf{27.8}$^\ddagger$&9.3$^\ddagger$&\textbf{32.1}$\pm 1.0^\star$ &\textbf{30.0}$^\dagger$   &\textbf{23.4}$^\ddagger$  &39.8$\pm 1.7$ \\  
				\hline 
			\end{tabular}%
			\begin{tablenotes}
				\footnotesize
				\item[1] We report the average performance of \textbf{10 runs on Charades-STA and DiDeMo, and 5 runs on ANet-Cap}, with different random seeds for network initialization. For each run, we select the best-performing model on original testing set to conduct two rounds of OOD testing, \textit{i.e.}, OOD-1 and OOD-2, with different values of $\rho$ mentioned in section \ref{OOD}. "-" means that the result on the dataset is not reported by the paper or its model is unavailable.
			\end{tablenotes}
	\end{threeparttable}}
	\vspace{-0.2in}
	\label{tab:SOTA}
\end{table*}

\subsubsection{\textbf{Metrics}}
We adopt the rank-1 accuracy (\textbf{R@1}) with different IoU thresholds (\textbf{IoU>$m$}), and \textbf{mIoU} as the evaluation metrics, following \cite{gao2017tall,zhang2019learning,zhang-etal-2020-span}. The "R@1, IoU>$m$" denotes the percentage of queries having top-1 retrieval result whose IoU score is larger than $m$. "mIoU" is the average IoU score of the top-1 results over all testing queries. In our experiments, we use $m\in\{0.5,0.7\}$ for ANet-Cap and Charades-STA, and $m\in\{0.7, 1.0\}$ for DiDeMo. Different from existing efforts which also report "R@5" accuracy, we only report {"R@1, IoU>$m$" ($m\geq 0.5$) and "mIoU"} for stricter evaluation.
\subsubsection{\textbf{Out-of-Distribution Testing}}\label{OOD}
The split of datasets in Table \ref{dataset_statistics} follows the Independent and Identically Distributed (IID) assumption, which is insufficient to evaluate the generalization. Instead, we evaluate the models not only using IID testing, 
 but also adopting {Out-of-Distribution (OOD)} testing~\cite{teney2020value}. The idea is to insert a randomly-generated video clip with $\rho$ seconds at the beginning of videos. Then the temporal length of each video is changed to $\tau + \rho$ and the timestamps of target moment are changed accordingly, \textit{i.e.}, $(\tau_s + \rho, \tau_e + \rho)$. For each dataset, we select the best-performing model on original testing set to conduct \textbf{two rounds of OOD testing} with different $\rho$. We use $\rho\in\{30,60\}$ for ANet-Cap, $\rho\in\{10,15\}$ for Charades-STA, and $\rho\in\{15,30\}$ for DiDeMo. The frame feature of the inserted video clip is randomly generated from a normal distribution. By the OOD testing, we keep the matching relationship unchanged while significantly changing the location distribution of moments to test the generalizability of models. 
\subsubsection{\textbf{Comparison Methods}} (1) \textit{Baselines}. As mentioned in section \ref{dcm_imp}, we implement our method with two matching networks: \textbf{CMI} and \textbf{TCN}. We use them as two baseline methods by just replacing $\bar{\mathbf{v}}+\mathbb{E}_{\bm{\ell}}\left[{h}_{qv}(\bm{\ell})\right]$ with original representation $\mathbf{v}$. Our methods are termed by adding a suffix: \textbf{+DCM}. The kernel size of the convolutional operation in CMI is $3\times 3$ and TCN is implemented by a 3-layer CNN with kernel size $5\times 5$. We also include two \textbf{biases-based methods} that explicitly exploit the temporal location biases: a) \textbf{Freq-Prior}: it models the moment frequency prior $P(Y|L)$ where the prediction on each moment candidate only depends on the frequency of location annotations in training set; b) \textbf{Blind-TAN}~\cite{otani2020uncovering}: it models $P(Y|Q,L)$ that excludes moment representation from the model input, implemented by replacing the moment feature in TCN with the positional embedding~\cite{vaswani2017attention} of moment location $(t_s,t_e)$. (2) \textit{State-of-the-art methods (SOTAs)}. We also compare our methods with the reported results of SOTAs on IID testing sets, such as SCDM~\cite{NIPS2019_8344}, 2D-TAN~\cite{zhang2019learning}, DRN~\cite{zeng2020dense}, LGI~\cite{mun2020LGI}, and VSLNeT~\cite{zhang-etal-2020-span}. 
\subsubsection{\textbf{Implementation Details}}
For the language query, we use 300d glove vectors as the word embeddings. The dimension of query and moment representations is set to $d=512$. The batchsize is set to 64 for ANet-Cap and DiDeMo, and 32 for Charades-STA, respectively. We adopt Adam as the optimizer with learning rate $1e-4$. The hyperparameters $\lambda_1$ and $\lambda_2$ are fixed as 1 and 0.001, respectively. We use the PCA-reduced C3D~\cite{tran2015learning} features as frame-level representation of ANet-Cap. We use three widely-used backbone networks (C3D, I3D~\cite{carreira2017quo}, and VGG~\cite{VGG}) to extract video representation of Charades-STA, respectively. We use the flow features in \cite{anne2017localizing} as video representation of DiDeMo. We follow \cite{zhang2019learning} to sample moment candidates from video and set the number of clips $T$=16 for ANet-Cap and Charades-STA and $T$=6 for DiDeMo. 
\textit{Besides, we exclude the long moments ($L_\textrm{mom}/L_\textrm{vid}\geq0.5$, nearly 4k in testing set) for OOD testing on ANet-Cap, since we found the OOD performance of Freq-Prior on these long moments is quite high}. 

\subsection{Overall Performance Comparison}
\subsubsection{\textbf{Comparison with Baselines}} \label{baselien_comp}
We report the empirical results of all baseline methods in Table \ref{tab:SOTA} on both IID (\textit{i.e.}, original) and OOD testing sets. The relative improvement and statistical significance test are performed between DCM methods and their baseline counterparts. We have the following observations:
\begin{itemize}
	\item All three datasets have strong temporal location biases, revealed by the two biased methods. By using the prior $P(Y|L)$ without training, Freq-Prior reports nearly 30\% mIoU on all IID testings. By using $P(Y|Q,L)$, Blind-TAN reports much higher R@1(IoU>0.5) accuracy. In particular, a SOTA result (45.3\%) is reported in ANet-Cap. When evaluated by OOD testing, their performances drop significantly. The results support the empirical analysis in \cite{otani2020uncovering} and indicate the necessity of OOD testing.
	\item The DCM methods consistently improve TCN and CMI in all settings, which suggests that DCM is agnostic to the methods, datasets, backbones, and distributions of testing sets. In particularly, the improvements on OOD testing are typically larger than those on IID testing. For example, the relative improvement \textit{w.r.t.} R@1(IoU>0.5) is 2.82\% on IID testing and 23.68\% on OOD testing on ANet-Cap. It reflects that traditional VMR models are vulnerable to the biases in datasets and sensitive to the distribution changes of moment annotations, and clearly demonstrates the high effectiveness of our DCM. We attribute such improvement to the following aspects: 1) the disentangling module reduces the effect of location on moment representation in the feature level, endowing the models with a stronger ability of precisely exploiting the video content; and 2) by the causal intervention, DCM is forced to fairly consider all the possible locations of target moment, via backdoor adjustment, to make a robust prediction which avoids exploiting the temporal location biases in datasets and therefore improves the generalizability of models.
	\item We find that the improvements of DCM on Charades-STA are much larger than those on the other datasets. It might be due to its relatively small training set, as shown in Table \ref{dataset_statistics}. Models can be more easily misled by the temporal location biases  when trained with less data. In particular, the TCN method, with more training parameters, reports much worse generalization performance than the lightweight method, CMI, on the small dataset.
	\item Even on the two OOD testing sets, Blind-TAN still reports more than 10\% R@1(IoU>0.5) score on ANet-Cap and Charades-STA, which indicates the high correlation between query and location. In fact, the prior $P(Y|Q,L)$ can be mitigated by re-split the dataset based on moment locations, \textit{i.e.}, forcing the locations of target moments in testing set to be unseen in training set. For easy comparison with existing results, we do not re-split the datasets.
	\item We find the OOD evaluation results on ANet-Cap are much lower than those of other datasets, since we exclude the long moments from its ODD testing. The reason is that moment length also leads to a prior in $P(Y|L)$. For the long target moments with length $L_\textrm{mom}/L_\textrm{vid}\geq0.5$, we can obtain 100\% R@1(IoU>0.5) by just returning the whole video as the prediction. The number of long moments in the testing set of ANet-Cap is nearly 4K. We remove them to keep the effectiveness of OOD testing.
\end{itemize}

\subsubsection{\textbf{Comparison with State-of-the-art Results}} We include the reported results of recent SOTA methods, such as LGI~\cite{mun2020LGI} and VSLNet~\cite{zhang-etal-2020-span}, in Table \ref{tab:SOTA}. We observe that TCN+DCM and CMI+DCM achieve new SOTA results \textit{w.r.t.} R@1(IoU>0.7) on IID testing: 27.7\% on ANet-Cap, 27.8\% on DiDeMo, and {34.4\%, 37.8\%, 28\%} on Charades-STA with different backbones. We also report the OOD testing results of LGI and VSLNet using their released models or codes. We find that our DCM methods significantly outperform SOTA methods \textit{w.r.t.} R@1(IoU>0.7) on OOD testing, \textit{e.g.}, 27.4\% relative improvement on OOD-1 on ANet-Cap and 26.7\% improvement on OOD-2 on Charades-STA (I3D) over LGI. The result is consistent with the comparison with baseline counterparts. It demonstrates that DCM can not only retain the good location bias to improve the accuracy in original testing set, but also remove the bad confounding effects of location to improve the generalization.

\begin{table}[tb]
	\renewcommand{\arraystretch}{0.95}
	\renewcommand\tabcolsep{3pt}
	\definecolor{white}{rgb}{1,1,1}
	\caption{Ablation studies. The $\ast$ and $\star$ denote the statistical significance for $p<0.05$ and $p<0.01$, respectively.}
	\vspace{-0.12in}
	\centering
	\scalebox{0.9}{%
		\begin{threeparttable}		
			\begin{tabular}{c|l||ll||ll||ll} 
				\hline
				&\multirow{3}{*}{\textbf{TCN+DCM}} &
				\multicolumn{2}{c||}{\textbf{ANet-Cap}}&\multicolumn{2}{c||}{\textbf{Charades.}}&\multicolumn{2}{c}{\textbf{DiDeMo}}\\
				\cline{3-8} 
				&&\multicolumn{2}{c||}{\textbf{IoU>m}} 
				&\multicolumn{2}{c||}{\textbf{IoU>m}} 
				&\multicolumn{2}{c}{\textbf{IoU>m}} \\			
				\cline{3-8}		
				&	& 0.5   & 0.7  & 0.5   & 0.7   & 0.7   & 1.0 \\ 
				\cline{1-8}		
				&\textbf{Full Model}&\underline{44.9}&\underline{27.7}&\underline{59.7}&\underline{\textbf{37.8}} &\underline{\textbf{37.5}}&\underline{\textbf{27.6}} \\ 
				\cline{2-8}		
				&\textit{w/o feat. disent.}&43.7$^\star$&26.7$^\star$&\textit{55.0}$^\star$&\textit{33.7}$^\star$ &36.2$^\star$&26.7$^\star$\\ 
				&\textit{w/o indep. loss}&44.7&27.7&\textbf{59.8}&37.4 &37.1$^\star$&27.3 \\ 
				&\textit{w/o recon. loss}&\textit{42.7}$^\star$ &\textit{25.7}$^\star$&55.8$^\star$&34.5$^\star$ &\textit{35.7}$^\star$&\textit{26.6}$^\star$ \\ 
				&\textit{w/o loc. feat.}&{44.4}$^\ast$ &{27.1}$^\ast$&\textbf{59.8}&37.3$^\ast$ &37.1$^\star$&27.4$^\ast$ \\ 
				&\textit{w/o counterf. loss}&45.3&28.9$^\star$&58.2$^\star$&37.3$^\star$ &37.0$^\star$&26.7$^\star$\\ 		
				\multirow{-6}{*}{\rotatebox{90}{\textbf{IID Test}}}&\textit{w/o causal interv.}&\textbf{45.6}&\textbf{29.5}&57.7$^\star$&37.1$^\star$ &36.8$^\star$&26.9$^\star$ \\ 
				\hline
				\hline
				&\textbf{Full Model}&\underline{\textbf{18.2}}&\underline{\textbf{7.9}}&\underline{\textbf{44.4}}&\underline{\textbf{19.7}} &\underline{\textbf{31.6}}&\underline{\textbf{24.4}}\\ 
				\cline{2-8}		
				&\textit{w/o feat. disent.}&17.7 &7.3$^\ast$&34.3$^\star$&17.0$^\star$ &{29.3}$^\star$&{22.4}$^\star$ \\ 
				&\textit{w/o indep. loss}&18.1 &7.7&43.8&19.1 &31.0&23.9 \\ 
				&\textit{w/o recon. loss}&17.8&7.2&38.2$^\star$&18.6 &30.7&22.7$^\star$ \\ 
				&\textit{w/o loc. feat.}&{17.7} &{7.1}$^\ast$&43.5&19.2 &\textit{28.4}$^\star$&\textit{22.3}$^\star$ \\ 
				&\textit{w/o counterf. loss}&17.0$^\ast$ &7.3&40.2$^\star$&18.0$^\star$ &31.1&23.5 \\ 		
				\multirow{-6}{*}{\rotatebox{90}{\textbf{OOD-1 Test}}}	&\textit{w/o causal interv.}&\textit{16.4}&\textit{6.7}$^\star$&\textit{34.1}$^\star$&\textit{15.9}$^\star$ &31.1&23.3$^\star$ \\ 					
				\hline 
			\end{tabular}
	\end{threeparttable}}
	\vspace{-0.14in}
	\label{tab:abl}
\end{table}

\subsection{Study of DCM}\label{study}
For a better understanding of DCM, we use TCN+DCM as an example to analyze the contribution of individual components. Empirical results on the three datasets\footnote{If not specifically stated, we use the I3D backbone for Charades-STA by default.} are shown in Table \ref{tab:abl} and Figure \ref{fig4}.

\subsubsection{\textbf{Effect of Feature Disentangling}} 
\begin{figure}[tb]
	\centering
	\includegraphics[width=3.1in]{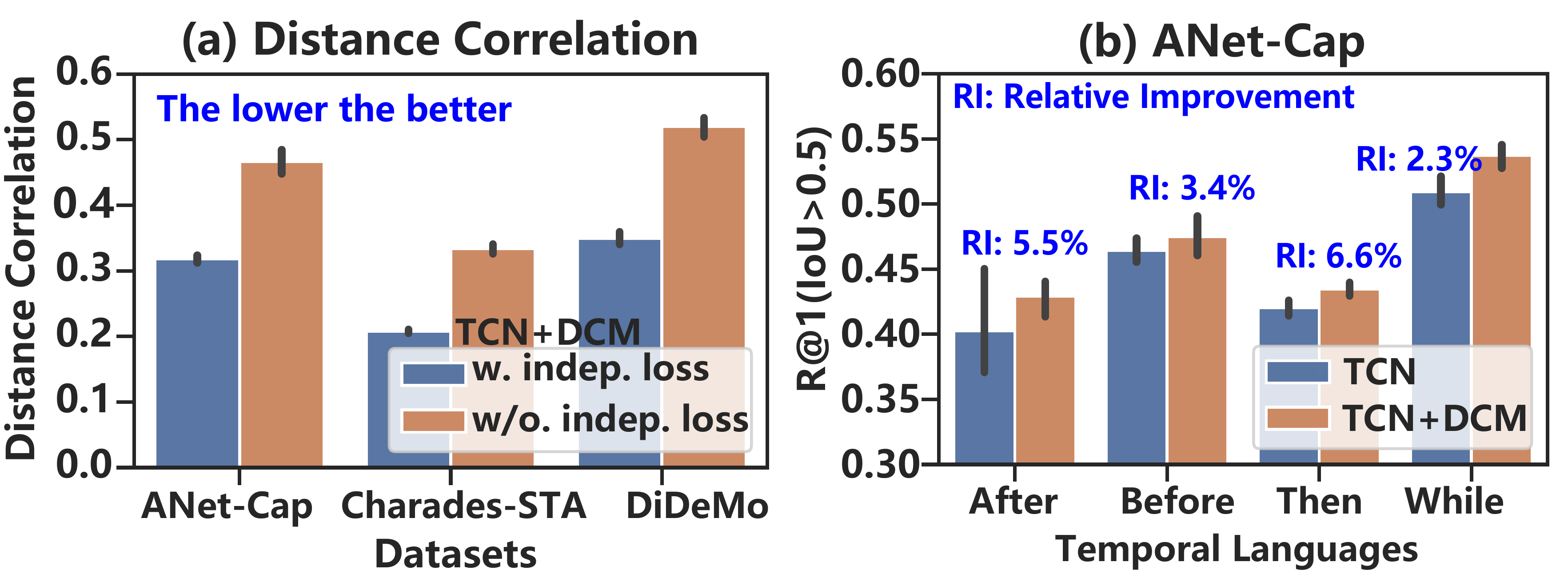}
	\vspace{-0.12in}
	\caption{(a) Distance correlation of TCN+DCM on 3 datasets; (b) Performance comparison (R@1, IID) on 4 query subsets of ANet-Cap with specific temporal languages.}
	\vspace{-0.2in}
	\label{fig4}
\end{figure}

As a core of DCM, the disentangling module decomposes moment representation into two latent factors: content and location using two specific losses, which ensures the visual content of moments be fairly treated and perceived. Here, we investigate, in Table \ref{tab:abl}, how this module affect the performance by alternatively removing the whole module (\textit{w/o feat. disent.}), the independence loss (\textit{w/o indep. loss}), the reconstruction loss (\textit{w/o recon. loss}) , and the disentangled location feature $\bm{\ell}^v$ in $\bar{\mathbf{v}}$ (\textit{w/o loc. feat.}) from the model. We have several findings:
\begin{itemize}
	\item We observe clear performance drop in all settings when removing the disentangling module. The reasons are two folds: 1) The entangled moment feature is weak in capturing the core feature of moment content. The latent location factor may dominate the moment representation, leading to poor generalization; and 2) without the disentangling, the confounder set has to be composed by the positional embedding of original locations $(\tau_s, \tau_e)$. As mentioned in section \ref{disentangled}, injecting the extra location feature into $\mathbf{v}$ may further exacerbate the correlation between $V$ and $L$, which hurts the model expressiveness. The results support the analysis in \cite{denton2017unsupervised} on the importance of  video representation disentangling and clearly indicate the necessity of the module in DCM.
	\item We use distance correlation to model the independence between the moment content and location vectors. The comparisons, \textit{w.r.t.} distance correlation and R@1 accuracy, are reported in Figure \ref{fig4} (a) and Table \ref{tab:abl}, respectively. We observe that a better OOD performance is substantially coupled with a higher independence across the board. This also supports the correlation between the performance and feature disentangling mentioned above. \textit{w/o indep. loss} hurts less the IID performance, since we use a small $\lambda_2=0.001$ to modulate its effect for stable training.
	\item  \textit{w/o recon. loss} leads to consistently worse IID and OOD results. The reason is that the latent location factor is not disentangled accurately, and then causal intervention fails to cut off the backdoor path. The result is consistent with the above analyses.
	\item In our implementation of DCM, we additionally keep the disentangled location feature $\bm{\ell}^v$ in the moment representation $\bar{\mathbf{v}}$ to slightly improve the weight of current location. We observe that \textit{w/o loc. feat.} leads to clear performance drop in OOD testing. The results suggest the necessity of keeping the good location bias. The location feature facilitates the temporal context modeling, which is especially crucial to handle the queries including temporal languages, such \textit{before} or \textit{after}. That can be supported by the performance comparison across 4 query subsets with different temporal languages in Figure \ref{fig4} (b). We find that, compared with the overall comparison (44.9\% \textit{vs.} 43.3\% \textit{w.r.t.} R@1(IoU>0.5)) between TCN+DCM and TCN on ANet-Cap, the improvements in Figure \ref{fig4} (b) are more significant. It further verifies the advantages of DCM in exploiting the good location bias.
\end{itemize}

\subsubsection{\textbf{Effect of Causal Intervention}} Causal Intervention is the key module in DCM. It intervenes the multimodal input based on backdoor adjustment to cut off the backdoor path. It consists of two components: 1) representation intervention, \textit{i.e.}, the addition of $\mathbb{E}_{\bm{\ell}}\left[{h}_{qv}(\bm{\ell})\right]$ to $\bar{\mathbf{v}}$; and 2) counterfactual loss $\mathcal{L}^{-}_{bce}$ that endows DCM with the ability of counterfactual thinking. We investigate its effect in Table \ref{tab:abl} by alternatively removing the whole module (\textit{w/o causal interv.}) and the counterfactual loss (\textit{w/o counterf. loss}) from the model. We have the following observations:
\begin{itemize}
	\item \textit{w/o causal interv.} leads to significant performance drops in OOD testing results. Compared with the full model, the relative performance drops \textit{w.r.t.} R@1(IoU>0.7) are 15.2\%, 19.3\%, and 4.5\% on three datasets, respectively. In particular, we surprisingly observe a large IID improvement (27.7$\rightarrow$29.5) but a significant OOD drop (7.9$\rightarrow$6.7) on ANet-Cap (the largest dataset) when removing causal intervention. This suggests that, without causal intervention, the model is at higher risk of exploiting dataset biases while making less use of video content. It is consistent with the remarkable performance of Blind-TAN on ANet-Cap in Table \ref{tab:SOTA}, which demonstrates that the issue of dataset biases is more serious in ANet-Cap. Overall, the results and analysis clearly indicate the effectiveness of causal intervention and further highlight the importance of introducing OOD testing into evaluation.
	\item If only removing the counterfactual loss $\mathcal{L}^{-}_{bce}$ from our learning objective in Eq. (\ref{full_loss}), we observe substantial performance drop of OOD testing results. This is consistent with the observation of \textit{w/o causal interv.}, which suggests that counterfactual loss is a perfect complement to the representation intervention in the loss level. It is able to prevent DCM from exploiting the location biases by penalizing high prediction scores on the biased locations in a negative video, like the prediction of Freq-Prior and Blind-TAN. 
\end{itemize}

\begin{figure}[tb]
	\centering
	\includegraphics[width=3in]{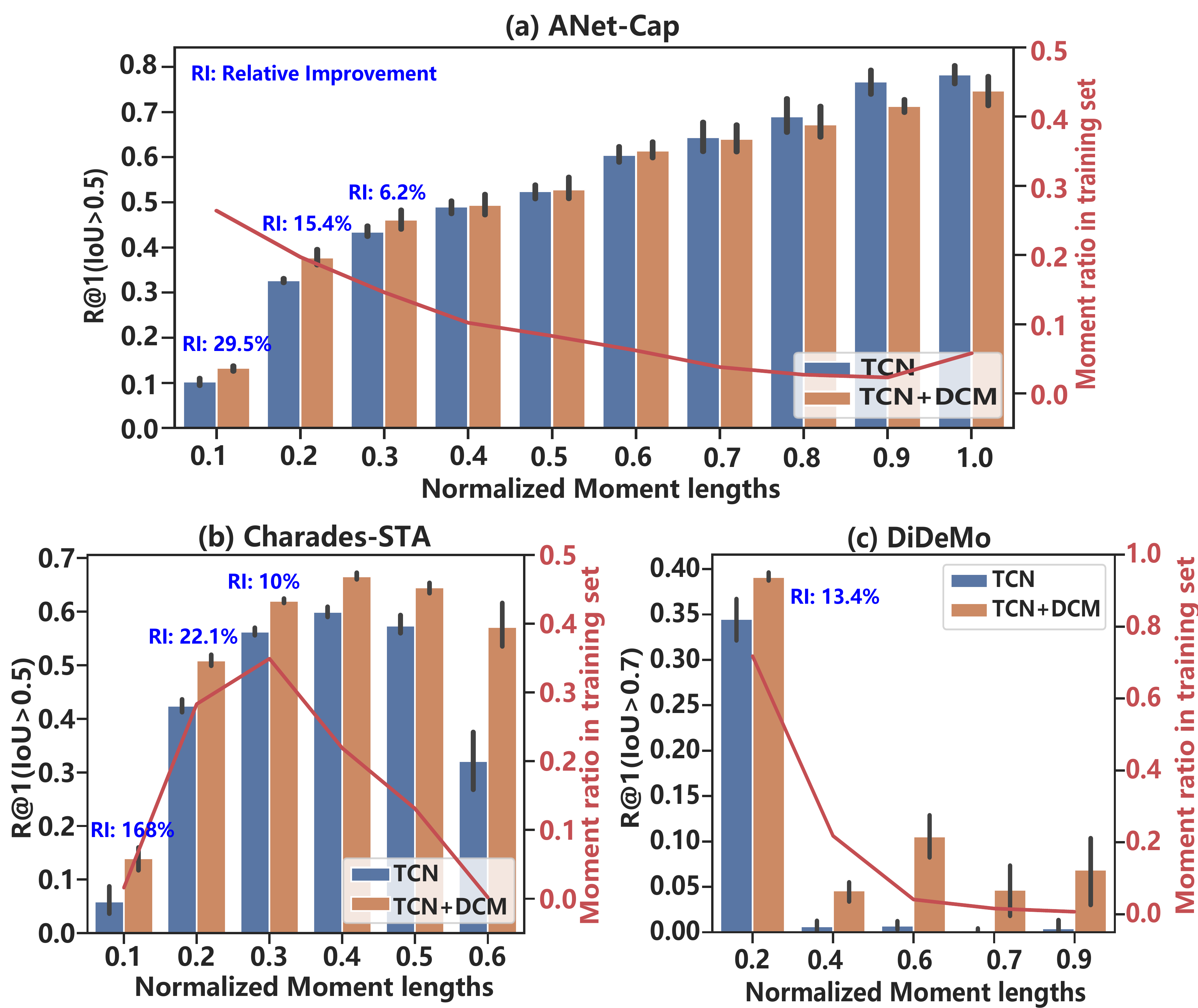}
	\vspace{-0.12in}
	\caption{Performance comparison (R@1, IID) \textit{w.r.t.} the temporal lengths of annotated moments in three datasets. The ratio of moment annotations in each group to the total number in training set is also plot.}
	\vspace{-0.2in}
	\label{fig5}
\end{figure}
\subsubsection{\textbf{On the Moment Length}} We investigate how DCM affect the IID performance \textit{w.r.t.} the moment length (\textit{the shorter the more challenging}).  
The empirical results are reported in Figure \ref{fig5}. We have the following findings:
\begin{itemize}
	\item We find the temporal lengths of annotated moments in ANet-Cap are quite diverse and there are nearly 25\% annotations with duration $L_\textrm{mom}/L_\textrm{vid}\geq0.5$. The baseline TCN achieves higher performance \textit{w.r.t.} R@1(IoU>0.5) on the long moment groups. The improvement might be attributed to the usage of dataset bias on moment lengths, as analyzed at the end of section \ref{baselien_comp}. The finding supports the observation in \cite{otani2020uncovering}. With causal intervention and counterfactual training, DCM-based method is prevented from exploiting such kind of dataset bias, thus reporting lower performance on these moment groups. While on the short moment groups in Figure \ref{fig5} (a), TCN+DCM reports significantly better performance over TCN. These results are consistent with the observation on ANet-Cap in Table \ref{tab:abl}.
	\item We observe substantial improvement of TCN+DCM over its counterpart, \textit{w.r.t.} R@1(IoU>0.5), across all groups in Charades-STA. In particular, on the first group in Figure \ref{fig5} (b) where the average duration of moments is just about 3s, we observe 168\% relative improvement of TCN+DCM over TCN. The finding clearly demonstrates the advantages of DCM-based methods in retrieving very short video moments. We also observe a bad performance of TCN on the long moment group (\textit{0.6}). It is not contrast to the observation from Figure \ref{fig5} (a), since there are just 0.2\% long moments with normalized length \textit{0.6} in the training set of Charades-STA and thus the model can not learn the length bias of long moments with such limited data.
	\item For the comparison on DiDeMo, we observe from Figure \ref{fig5} (c) that most of its annotations fall into the short moment group (\textit{0.2}). Our DCM method achieves significant improvement (13.4\%, \textit{w.r.t.} R@1(IoU>0.7)) over its counterpart on this group.
\end{itemize}
Overall, the observations from Figure \ref{fig5} clearly indicate the effectiveness of our DCM on moment retrieval in videos.
\begin{figure}[tb]
	\centering
	\includegraphics[width=3.1in]{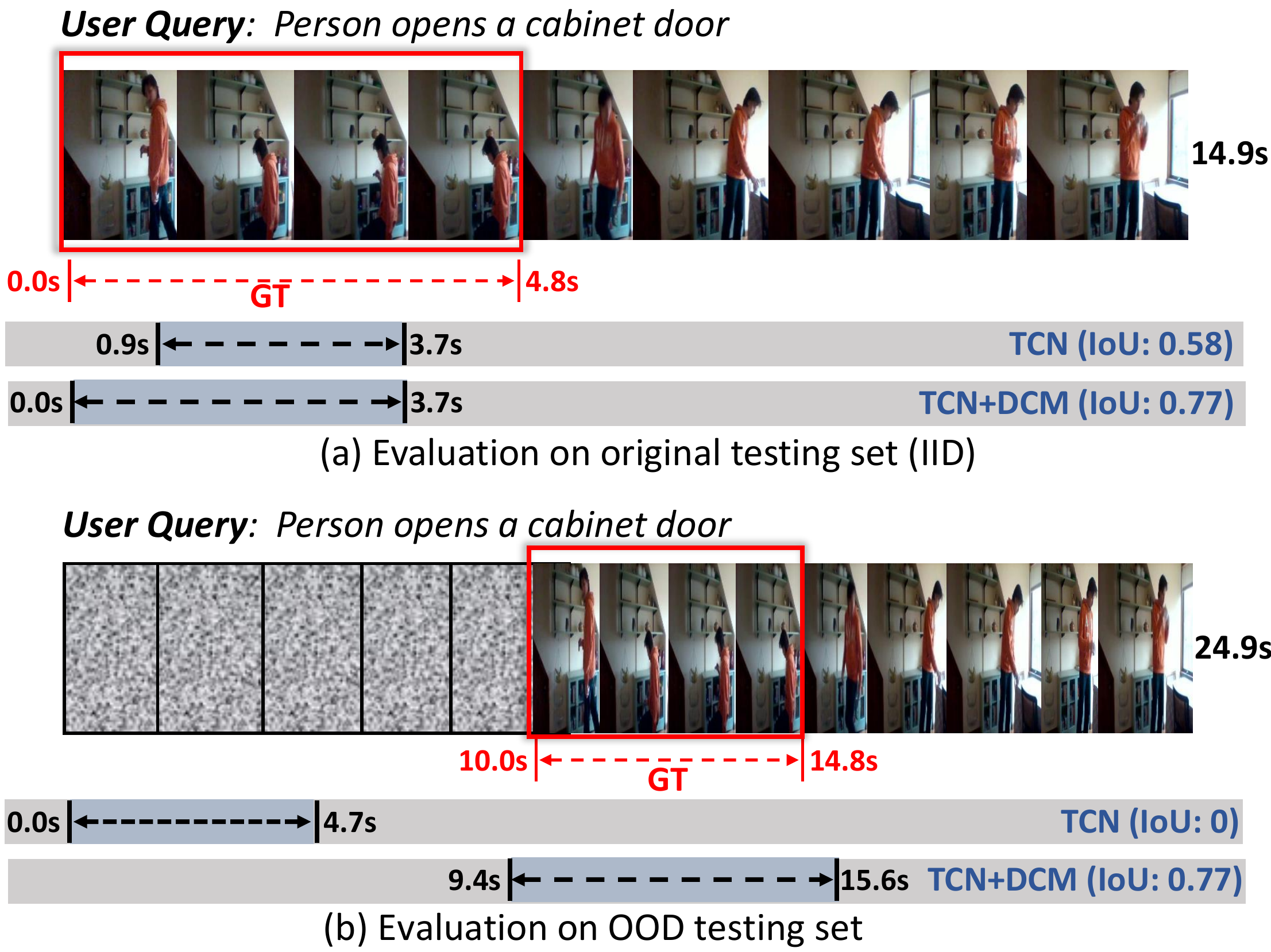}
	\vspace{-0.12in}
	\caption{The sampled moment retrieval results, \textit{w.r.t.} R@1, on both IID and OOD testing sets of Charades-STA. }
	\vspace{-0.2in}
	\label{fig6}
\end{figure}
\subsubsection{\textbf{Qualitative Results}}
In this section, we present some qualitative results, as shown in Figure \ref{fig6}, to give an intuitive impression of the effectiveness of our DCM for VMR.  Figure \ref{fig6} shows a real example from Charades-STA: retrieving the specific moment of "\textit{Person opens a cabinet door}" in the given video. We illustrate the R@1 retrieval results of TCN+DCM and its counterpart by both IID and OOD testing. We have the following findings:
\begin{itemize}
	\item For the query in Figure \ref{fig6}, TCN+DCM performs stably against the distribution changes of moment location. In particular, in Figure \ref{fig6} (a), its top-1 retrieval result is the moment at timestamps ($0.0s, 3.7s$), which has more than 70\% overlap with the groundtruth annotation. When moved to the OOD testing by inserting a randomly-generated video segment at the beginning of the video, TCN+DCM can still find the best candidate with high IoU score. It reflects that DCM is not only robust to the \textit{distribution changes} but also robust to the noisy temporal context.
	\item TCN is sensitive to the distribution changes. The IoU score of its top-1 retrieval result significantly drops from 58\% in Figure \ref{fig6} (a) to zero in (b). The result is consistent with the previous analysis. It indicates that the location biases have severely affected the training of TCN, leading to low robustness.
\end{itemize}

\section{Related Work}
\subsubsection{\textbf{Video Moment Retrieval}}
Video moment retrieval (VMR) \cite{gao2017tall,anne2017localizing}, also termed temporal grounding/localization, attracts increasing attention in recent years~\cite{chen2018temporally,hendricks2018localizing,NIPS2019_8344}. It aims to retrieve a short video segments in videos using language queries. The key to solving VMR is how to effectively learn the cross-modal matching relationship between the language and video segments.  Existing works mainly follow a cross-modal matching framework~\cite{anne2017localizing,gao2017tall,hendricks2018localizing,liu2018attentive,liu2018cross} that first generates a set of candidate moments, and then ranks them based on the matching score between the moment candidates and the language query. Gao \textit{et al.}~\cite{gao2017tall} and Liu \textit{et al.}~\cite{liu2018attentive,liu2018cross} used the traditional multi-scale sliding window to sample the candidates, which is simple but has low recall of short candidates. Zhang \textit{et al.}~\cite{zhang2019learning} proposed a stacked pooling network to densely enumerate all the valid candidates and store them in a three-dimensional tensor matrix to preserve the temporal structure, which allows CNNs to be applied in VMR for learning moment representation. The modeling of temporal structure of video moments, introduced in~\cite{zhang2019learning}, becomes increasingly popular~\cite{liu2020jointly,wang2020dual}.
The difference among existing efforts mainly lies in the design of multimodal fusion module, \textit{e.g.}, cross-modal feature fusion~\cite{liu2018attentive,mun2020LGI,zhang-etal-2020-span} or pairwise similarity learning~\cite{anne2017localizing} following metric learning framework~\cite{yang2017person}. The former is more popular.
Chen \textit{et al.}~\cite{chen2018temporally}, Yuan \textit{et al.}~\cite{yuan2019find}, and Mun \textit{et al.}~\cite{mun2020LGI} designed a complex visual-textual interaction/attention modules for cross-modal interaction. In particular, Mun \textit{et al.}~\cite{mun2020LGI} presented a video-text interaction algorithm in capturing relationships of semantic phrases and video segments by modeling local and global contexts, reporting SOTA performance. Yuan \textit{et al.} \cite{NIPS2019_8344} and Zhang \textit{et al.} \cite{zhang2019man,zhang2019learning} applied layer-wise temporal CNNs to model the temporal relations between moments, receiving more attention recently. \\
\noindent{\textbf{Difference between DCM and existing works:}} \textbf{1)} We focus on improving the generalizability of models, since recent findings~\cite{otani2020uncovering} reported that SOTA models tend to exploit dataset biases while agnostic to video content. Specifically, our DCM applies causal intervention to remove the confounding effects of moment location, which encourages the model to fairly consider all the possible locations of target for making a robust prediction; \textbf{2)} We introduce counterfactual training that endows the model with the ability of counterfactual thinking to answer the question "What if the target moment does not exist in the given video?". Different from our work, existing methods all assume that the target moment must be in the give video; \textbf{3)} We demonstrate the importance and necessity of applying OOD testing to evaluate the generalizability of VMR models, beyond the widely-used IID testing;  \textbf{4)} We empirically indicate that existing methods are vulnerable to the temporal location biases in datasets. Our DCM has the potential to be coupled with exiting methods to improve their generalizability.
\subsubsection{\textbf{Causal Inference}}
Recently, causal inference~\cite{pearl2016causal,pearl2018book} has attracted increasing attention in information retrieval and multimedia for removing dataset biases in domain-specific applications, such as recommendation~\cite{sato2020unbiased,wang2020causal,yap2007discovering,feng2021causalgcn,wang2021clickbait}, visual dialog~\cite{qi2020two}, segmentation~\cite{zhang2020causal}, unsupervised feature learning~\cite{wang2020visual}, video action localization~\cite{liu2021blessing}, and scene graph~\cite{zhang2020causal}, \textit{etc.} The general purpose of causal inference is to empower the models the ability of pursuing the causal effect, thus leading to more robust decision. In particular, Qi \textit{et al.}~\cite{qi2020two} presented causal principles to improve visual dialog, where causal intervention is used to remove the effect of an unobserved confounder. 
Zhang \textit{et al.}~\cite{zhang2020causal} introduced causal inference into weakly-supervised semantic segmentation by removing confounding bias of context prior. Wang \textit{et al.}~\cite{wang2020visual} proposed to use causal intervention to learn "sense-making" visual knowledge beyond traditional visual co-occurrences. \textbf{Different from existing work}, we make a new attempt of causal inference~\cite{pearl2016causal} to solve the task of VMR. Specifically, we reveal the causalities between different components in VMR and point out that the temporal location of a video moment is a confounder that spuriously correlates the model prediction and multimodal input. To remove the harmful confounding effects, we develop a deconfounding method, DCM, that first disentangles moment representation to learn the core feature of visual content and then intervenes the multimodal input based on backdoor adjustment. This is the first causality-based work that addresses the temporal location biases of VMR, which is significantly different from a new VMR work  \cite{nan2021interventional}.

\section{Conclusion}
This paper proposed a deconfounded cross-modal matching method for the VMR task. By introducing out-of-distribution testing into the model evaluation, we empirically demonstrated that existing VMR models are sensitive to the distribution changes of moment annotations, and found that the temporal location biases,  hidden in current datasets, may severely affect the model training and prediction. Our proposed deconfounding algorithm is effective in improving the generalizability of VMR models. It can not only eliminate the bad but also retain the good bias of moment temporal location in videos. Our method has the potential to be effectively integrated with most exiting methods to improve their generalizability. This work showed initial attempts toward robust video moment retrieval against the dataset biases. In the future, we will further explore the intervention strategy on the user query to ensure a comprehensive understanding of complex user intents. Besides, we will also apply our deconfounding approach on other location-sensitive tasks, \textit{e.g.}, visual grounding~\cite{hong2019learning,yang2020weakly} and temporal activity localization~\cite{liu2021blessing}, to mitigate their location biases in datasets.

\section{Acknowledgments}
This research/project is supported by the Sea-NExT Joint Lab.

\bibliographystyle{ACM-Reference-Format}
\bibliography{sample-base}


\begin{thebibliography}{52}


\ifx \showCODEN    \undefined \def \showCODEN     #1{\unskip}     \fi
\ifx \showDOI      \undefined \def \showDOI       #1{#1}\fi
\ifx \showISBNx    \undefined \def \showISBNx     #1{\unskip}     \fi
\ifx \showISBNxiii \undefined \def \showISBNxiii  #1{\unskip}     \fi
\ifx \showISSN     \undefined \def \showISSN      #1{\unskip}     \fi
\ifx \showLCCN     \undefined \def \showLCCN      #1{\unskip}     \fi
\ifx \shownote     \undefined \def \shownote      #1{#1}          \fi
\ifx \showarticletitle \undefined \def \showarticletitle #1{#1}   \fi
\ifx \showURL      \undefined \def \showURL       {\relax}        \fi
\providecommand\bibfield[2]{#2}
\providecommand\bibinfo[2]{#2}
\providecommand\natexlab[1]{#1}
\providecommand\showeprint[2][]{arXiv:#2}

\bibitem[\protect\citeauthoryear{Anne~Hendricks, Wang, Shechtman, Sivic,
  Darrell, and Russell}{Anne~Hendricks et~al\mbox{.}}{2017}]%
        {anne2017localizing}
\bibfield{author}{\bibinfo{person}{Lisa Anne~Hendricks},
  \bibinfo{person}{Oliver Wang}, \bibinfo{person}{Eli Shechtman},
  \bibinfo{person}{Josef Sivic}, \bibinfo{person}{Trevor Darrell}, {and}
  \bibinfo{person}{Bryan Russell}.} \bibinfo{year}{2017}\natexlab{}.
\newblock \showarticletitle{Localizing moments in video with natural language}.
  In \bibinfo{booktitle}{\emph{ICCV}}. \bibinfo{pages}{5803--5812}.
\newblock


\bibitem[\protect\citeauthoryear{Baldi and Sadowski}{Baldi and
  Sadowski}{2014}]%
        {baldi2014dropout}
\bibfield{author}{\bibinfo{person}{Pierre Baldi} {and} \bibinfo{person}{Peter
  Sadowski}.} \bibinfo{year}{2014}\natexlab{}.
\newblock \showarticletitle{The dropout learning algorithm}.
\newblock \bibinfo{journal}{\emph{Artificial intelligence}}
  \bibinfo{volume}{210} (\bibinfo{year}{2014}), \bibinfo{pages}{78--122}.
\newblock


\bibitem[\protect\citeauthoryear{Carreira and Zisserman}{Carreira and
  Zisserman}{2017}]%
        {carreira2017quo}
\bibfield{author}{\bibinfo{person}{Joao Carreira} {and} \bibinfo{person}{Andrew
  Zisserman}.} \bibinfo{year}{2017}\natexlab{}.
\newblock \showarticletitle{Quo vadis, action recognition? a new model and the
  kinetics dataset}. In \bibinfo{booktitle}{\emph{CVPR}}.
  \bibinfo{pages}{6299--6308}.
\newblock


\bibitem[\protect\citeauthoryear{Chen, Chen, Ma, Jie, and Chua}{Chen
  et~al\mbox{.}}{2018}]%
        {chen2018temporally}
\bibfield{author}{\bibinfo{person}{Jingyuan Chen}, \bibinfo{person}{Xinpeng
  Chen}, \bibinfo{person}{Lin Ma}, \bibinfo{person}{Zequn Jie}, {and}
  \bibinfo{person}{Tat-Seng Chua}.} \bibinfo{year}{2018}\natexlab{}.
\newblock \showarticletitle{Temporally grounding natural sentence in video}. In
  \bibinfo{booktitle}{\emph{EMNLP}}. \bibinfo{pages}{162--171}.
\newblock


\bibitem[\protect\citeauthoryear{Denton et~al\mbox{.}}{Denton
  et~al\mbox{.}}{2017}]%
        {denton2017unsupervised}
\bibfield{author}{\bibinfo{person}{Emily~L Denton} {et~al\mbox{.}}}
  \bibinfo{year}{2017}\natexlab{}.
\newblock \showarticletitle{Unsupervised learning of disentangled
  representations from video}. In \bibinfo{booktitle}{\emph{NeurIPS}}.
  \bibinfo{pages}{4414--4423}.
\newblock


\bibitem[\protect\citeauthoryear{Dong, Li, and Snoek}{Dong
  et~al\mbox{.}}{2018}]%
        {dong2018predicting}
\bibfield{author}{\bibinfo{person}{Jianfeng Dong}, \bibinfo{person}{Xirong Li},
  {and} \bibinfo{person}{Cees~GM Snoek}.} \bibinfo{year}{2018}\natexlab{}.
\newblock \showarticletitle{Predicting visual features from text for image and
  video caption retrieval}.
\newblock \bibinfo{journal}{\emph{IEEE Transactions on Multimedia}}
  \bibinfo{volume}{20}, \bibinfo{number}{12} (\bibinfo{year}{2018}),
  \bibinfo{pages}{3377--3388}.
\newblock


\bibitem[\protect\citeauthoryear{Dong, Li, Xu, Yang, Yang, Wang, and Wang}{Dong
  et~al\mbox{.}}{2021}]%
        {dong2021dual}
\bibfield{author}{\bibinfo{person}{Jianfeng Dong}, \bibinfo{person}{Xirong Li},
  \bibinfo{person}{Chaoxi Xu}, \bibinfo{person}{Xun Yang},
  \bibinfo{person}{Gang Yang}, \bibinfo{person}{Xun Wang}, {and}
  \bibinfo{person}{Meng Wang}.} \bibinfo{year}{2021}\natexlab{}.
\newblock \showarticletitle{Dual encoding for video retrieval by text}.
\newblock \bibinfo{journal}{\emph{IEEE Transactions on Pattern Analysis and
  Machine Intelligence}} (\bibinfo{year}{2021}).
\newblock


\bibitem[\protect\citeauthoryear{Feng, Huang, He, Xin, Wang, and Chua}{Feng
  et~al\mbox{.}}{2021}]%
        {feng2021causalgcn}
\bibfield{author}{\bibinfo{person}{Fuli Feng}, \bibinfo{person}{Weiran Huang},
  \bibinfo{person}{Xiangnan He}, \bibinfo{person}{Xin Xin},
  \bibinfo{person}{Qifan Wang}, {and} \bibinfo{person}{Tat-Seng Chua}.}
  \bibinfo{year}{2021}\natexlab{}.
\newblock \showarticletitle{Should Graph Convolution Trust Neighbors? A Simple
  Causal Inference Method}. In \bibinfo{booktitle}{\emph{SIGIR}}.
\newblock


\bibitem[\protect\citeauthoryear{Gao, Sun, Yang, and Nevatia}{Gao
  et~al\mbox{.}}{2017}]%
        {gao2017tall}
\bibfield{author}{\bibinfo{person}{Jiyang Gao}, \bibinfo{person}{Chen Sun},
  \bibinfo{person}{Zhenheng Yang}, {and} \bibinfo{person}{Ram Nevatia}.}
  \bibinfo{year}{2017}\natexlab{}.
\newblock \showarticletitle{Tall: Temporal activity localization via language
  query}. In \bibinfo{booktitle}{\emph{ICCV}}. \bibinfo{pages}{5267--5275}.
\newblock


\bibitem[\protect\citeauthoryear{Hendricks, Wang, Shechtman, Sivic, Darrell,
  and Russell}{Hendricks et~al\mbox{.}}{2018}]%
        {hendricks2018localizing}
\bibfield{author}{\bibinfo{person}{Lisa~Anne Hendricks},
  \bibinfo{person}{Oliver Wang}, \bibinfo{person}{Eli Shechtman},
  \bibinfo{person}{Josef Sivic}, \bibinfo{person}{Trevor Darrell}, {and}
  \bibinfo{person}{Bryan Russell}.} \bibinfo{year}{2018}\natexlab{}.
\newblock \showarticletitle{Localizing Moments in Video with Temporal
  Language}. In \bibinfo{booktitle}{\emph{EMNLP}}. \bibinfo{pages}{1380--1390}.
\newblock


\bibitem[\protect\citeauthoryear{Hong, Liu, Mo, He, and Zhang}{Hong
  et~al\mbox{.}}{2019}]%
        {hong2019learning}
\bibfield{author}{\bibinfo{person}{Richang Hong}, \bibinfo{person}{Daqing Liu},
  \bibinfo{person}{Xiaoyu Mo}, \bibinfo{person}{Xiangnan He}, {and}
  \bibinfo{person}{Hanwang Zhang}.} \bibinfo{year}{2019}\natexlab{}.
\newblock \showarticletitle{Learning to compose and reason with language tree
  structures for visual grounding}.
\newblock \bibinfo{journal}{\emph{IEEE transactions on pattern analysis and
  machine intelligence}} (\bibinfo{year}{2019}).
\newblock


\bibitem[\protect\citeauthoryear{Hong, Yang, Wang, and Hua}{Hong
  et~al\mbox{.}}{2015}]%
        {hong2015learning}
\bibfield{author}{\bibinfo{person}{Richang Hong}, \bibinfo{person}{Yang Yang},
  \bibinfo{person}{Meng Wang}, {and} \bibinfo{person}{Xian-Sheng Hua}.}
  \bibinfo{year}{2015}\natexlab{}.
\newblock \showarticletitle{Learning visual semantic relationships for
  efficient visual retrieval}.
\newblock \bibinfo{journal}{\emph{IEEE Transactions on Big Data}}
  \bibinfo{volume}{1}, \bibinfo{number}{4} (\bibinfo{year}{2015}),
  \bibinfo{pages}{152--161}.
\newblock


\bibitem[\protect\citeauthoryear{JeffreyPennington and
  Manning}{JeffreyPennington and Manning}{2014}]%
        {jeffreypennington2014glove}
\bibfield{author}{\bibinfo{person}{RichardSocher JeffreyPennington} {and}
  \bibinfo{person}{ChristopherD Manning}.} \bibinfo{year}{2014}\natexlab{}.
\newblock \showarticletitle{Glove: Global vectors for word representation}. In
  \bibinfo{booktitle}{\emph{EMNLP}}.
\newblock


\bibitem[\protect\citeauthoryear{Karen~Simonyan}{Karen~Simonyan}{2015}]%
        {VGG}
\bibfield{author}{\bibinfo{person}{Andrew~Zisserman Karen~Simonyan}.}
  \bibinfo{year}{2015}\natexlab{}.
\newblock \showarticletitle{Very Deep Convolutional Networks for Large-Scale
  Image Recognition}. In \bibinfo{booktitle}{\emph{ICLR}}.
\newblock


\bibitem[\protect\citeauthoryear{Krishna, Hata, Ren, Fei-Fei, and
  Carlos~Niebles}{Krishna et~al\mbox{.}}{2017}]%
        {krishna2017dense}
\bibfield{author}{\bibinfo{person}{Ranjay Krishna}, \bibinfo{person}{Kenji
  Hata}, \bibinfo{person}{Frederic Ren}, \bibinfo{person}{Li Fei-Fei}, {and}
  \bibinfo{person}{Juan Carlos~Niebles}.} \bibinfo{year}{2017}\natexlab{}.
\newblock \showarticletitle{Dense-captioning events in videos}. In
  \bibinfo{booktitle}{\emph{ICCV}}. \bibinfo{pages}{706--715}.
\newblock


\bibitem[\protect\citeauthoryear{Liu, Qu, Liu, Dong, Zhou, and Xu}{Liu
  et~al\mbox{.}}{2020}]%
        {liu2020jointly}
\bibfield{author}{\bibinfo{person}{Daizong Liu}, \bibinfo{person}{Xiaoye Qu},
  \bibinfo{person}{Xiao-Yang Liu}, \bibinfo{person}{Jianfeng Dong},
  \bibinfo{person}{Pan Zhou}, {and} \bibinfo{person}{Zichuan Xu}.}
  \bibinfo{year}{2020}\natexlab{}.
\newblock \showarticletitle{Jointly Cross-and Self-Modal Graph Attention
  Network for Query-Based Moment Localization}. In
  \bibinfo{booktitle}{\emph{ACM MM}}. \bibinfo{pages}{4070--4078}.
\newblock


\bibitem[\protect\citeauthoryear{Liu, Wang, Nie, He, Chen, and Chua}{Liu
  et~al\mbox{.}}{2018a}]%
        {liu2018attentive}
\bibfield{author}{\bibinfo{person}{Meng Liu}, \bibinfo{person}{Xiang Wang},
  \bibinfo{person}{Liqiang Nie}, \bibinfo{person}{Xiangnan He},
  \bibinfo{person}{Baoquan Chen}, {and} \bibinfo{person}{Tat-Seng Chua}.}
  \bibinfo{year}{2018}\natexlab{a}.
\newblock \showarticletitle{Attentive moment retrieval in videos}. In
  \bibinfo{booktitle}{\emph{SIGIR}}. ACM, \bibinfo{pages}{15--24}.
\newblock


\bibitem[\protect\citeauthoryear{Liu, Wang, Nie, Tian, Chen, and Chua}{Liu
  et~al\mbox{.}}{2018b}]%
        {liu2018cross}
\bibfield{author}{\bibinfo{person}{Meng Liu}, \bibinfo{person}{Xiang Wang},
  \bibinfo{person}{Liqiang Nie}, \bibinfo{person}{Qi Tian},
  \bibinfo{person}{Baoquan Chen}, {and} \bibinfo{person}{Tat-Seng Chua}.}
  \bibinfo{year}{2018}\natexlab{b}.
\newblock \showarticletitle{Cross-modal moment localization in videos}. In
  \bibinfo{booktitle}{\emph{ACM MM}}. \bibinfo{pages}{843--851}.
\newblock


\bibitem[\protect\citeauthoryear{Liu, Chen, Chen, Deng, Huang, and Zhang}{Liu
  et~al\mbox{.}}{2021}]%
        {liu2021blessing}
\bibfield{author}{\bibinfo{person}{Yuan Liu}, \bibinfo{person}{Jingyuan Chen},
  \bibinfo{person}{Zhenfang Chen}, \bibinfo{person}{Bing Deng},
  \bibinfo{person}{Jianqiang Huang}, {and} \bibinfo{person}{Hanwang Zhang}.}
  \bibinfo{year}{2021}\natexlab{}.
\newblock \showarticletitle{The Blessings of Unlabeled Background in Untrimmed
  Videos}. In \bibinfo{booktitle}{\emph{CVPR}}.
\newblock


\bibitem[\protect\citeauthoryear{Mun, Cho, , and Han}{Mun
  et~al\mbox{.}}{2020}]%
        {mun2020LGI}
\bibfield{author}{\bibinfo{person}{Jonghwan Mun}, \bibinfo{person}{Minsu Cho},
  \bibinfo{person}{}, {and} \bibinfo{person}{Bohyung Han}.}
  \bibinfo{year}{2020}\natexlab{}.
\newblock \showarticletitle{{Local-Global Video-Text Interactions for Temporal
  Grounding}}. In \bibinfo{booktitle}{\emph{CVPR}}.
\newblock


\bibitem[\protect\citeauthoryear{Nan, Qiao, Xiao, Liu, Leng, Zhang, and Lu}{Nan
  et~al\mbox{.}}{2021}]%
        {nan2021interventional}
\bibfield{author}{\bibinfo{person}{Guoshun Nan}, \bibinfo{person}{Rui Qiao},
  \bibinfo{person}{Yao Xiao}, \bibinfo{person}{Jun Liu},
  \bibinfo{person}{Sicong Leng}, \bibinfo{person}{Hao Zhang}, {and}
  \bibinfo{person}{Wei Lu}.} \bibinfo{year}{2021}\natexlab{}.
\newblock \showarticletitle{Interventional Video Grounding with Dual
  Contrastive Learning}. In \bibinfo{booktitle}{\emph{CVPR}}.
\newblock


\bibitem[\protect\citeauthoryear{Nie, Yan, Wang, Hong, and Chua}{Nie
  et~al\mbox{.}}{2012}]%
        {nie2012harvesting}
\bibfield{author}{\bibinfo{person}{Liqiang Nie}, \bibinfo{person}{Shuicheng
  Yan}, \bibinfo{person}{Meng Wang}, \bibinfo{person}{Richang Hong}, {and}
  \bibinfo{person}{Tat-Seng Chua}.} \bibinfo{year}{2012}\natexlab{}.
\newblock \showarticletitle{Harvesting visual concepts for image search with
  complex queries}. In \bibinfo{booktitle}{\emph{ACM MM}}.
  \bibinfo{pages}{59--68}.
\newblock


\bibitem[\protect\citeauthoryear{Otani, Nakashima, Rahtu, and
  Heikkil{\"a}}{Otani et~al\mbox{.}}{2020}]%
        {otani2020uncovering}
\bibfield{author}{\bibinfo{person}{Mayu Otani}, \bibinfo{person}{Yuta
  Nakashima}, \bibinfo{person}{Esa Rahtu}, {and} \bibinfo{person}{Janne
  Heikkil{\"a}}.} \bibinfo{year}{2020}\natexlab{}.
\newblock \showarticletitle{Uncovering Hidden Challenges in Query-Based Video
  Moment Retrieval}. In \bibinfo{booktitle}{\emph{BMVC}}.
\newblock


\bibitem[\protect\citeauthoryear{Pearl, Glymour, and Jewell}{Pearl
  et~al\mbox{.}}{2016}]%
        {pearl2016causal}
\bibfield{author}{\bibinfo{person}{Judea Pearl}, \bibinfo{person}{Madelyn
  Glymour}, {and} \bibinfo{person}{Nicholas~P Jewell}.}
  \bibinfo{year}{2016}\natexlab{}.
\newblock \bibinfo{booktitle}{\emph{Causal inference in statistics: A primer}}.
\newblock \bibinfo{publisher}{John Wiley \& Sons}.
\newblock


\bibitem[\protect\citeauthoryear{Pearl and Mackenzie}{Pearl and
  Mackenzie}{2018}]%
        {pearl2018book}
\bibfield{author}{\bibinfo{person}{Judea Pearl} {and} \bibinfo{person}{Dana
  Mackenzie}.} \bibinfo{year}{2018}\natexlab{}.
\newblock \bibinfo{booktitle}{\emph{The book of why: the new science of cause
  and effect}}.
\newblock \bibinfo{publisher}{Basic Books}.
\newblock


\bibitem[\protect\citeauthoryear{Qi, Niu, Huang, and Zhang}{Qi
  et~al\mbox{.}}{2020}]%
        {qi2020two}
\bibfield{author}{\bibinfo{person}{Jiaxin Qi}, \bibinfo{person}{Yulei Niu},
  \bibinfo{person}{Jianqiang Huang}, {and} \bibinfo{person}{Hanwang Zhang}.}
  \bibinfo{year}{2020}\natexlab{}.
\newblock \showarticletitle{Two causal principles for improving visual dialog}.
  In \bibinfo{booktitle}{\emph{CVPR}}. \bibinfo{pages}{10860--10869}.
\newblock


\bibitem[\protect\citeauthoryear{Sato, Takemori, Singh, and Ohkuma}{Sato
  et~al\mbox{.}}{2020}]%
        {sato2020unbiased}
\bibfield{author}{\bibinfo{person}{Masahiro Sato}, \bibinfo{person}{Sho
  Takemori}, \bibinfo{person}{Janmajay Singh}, {and} \bibinfo{person}{Tomoko
  Ohkuma}.} \bibinfo{year}{2020}\natexlab{}.
\newblock \showarticletitle{Unbiased Learning for the Causal Effect of
  Recommendation}. In \bibinfo{booktitle}{\emph{Fourteenth ACM Conference on
  Recommender Systems}}. \bibinfo{pages}{378--387}.
\newblock


\bibitem[\protect\citeauthoryear{Sigurdsson, Varol, Wang, Farhadi, Laptev, and
  Gupta}{Sigurdsson et~al\mbox{.}}{2016}]%
        {sigurdsson2016hollywood}
\bibfield{author}{\bibinfo{person}{Gunnar~A Sigurdsson},
  \bibinfo{person}{G{\"u}l Varol}, \bibinfo{person}{Xiaolong Wang},
  \bibinfo{person}{Ali Farhadi}, \bibinfo{person}{Ivan Laptev}, {and}
  \bibinfo{person}{Abhinav Gupta}.} \bibinfo{year}{2016}\natexlab{}.
\newblock \showarticletitle{Hollywood in homes: Crowdsourcing data collection
  for activity understanding}. In \bibinfo{booktitle}{\emph{ECCV}}. Springer,
  \bibinfo{pages}{510--526}.
\newblock


\bibitem[\protect\citeauthoryear{Sz{\'e}kely, Rizzo, Bakirov,
  et~al\mbox{.}}{Sz{\'e}kely et~al\mbox{.}}{2007}]%
        {szekely2007measuring}
\bibfield{author}{\bibinfo{person}{G{\'a}bor~J Sz{\'e}kely},
  \bibinfo{person}{Maria~L Rizzo}, \bibinfo{person}{Nail~K Bakirov},
  {et~al\mbox{.}}} \bibinfo{year}{2007}\natexlab{}.
\newblock \showarticletitle{Measuring and testing dependence by correlation of
  distances}.
\newblock \bibinfo{journal}{\emph{The annals of statistics}}
  \bibinfo{volume}{35}, \bibinfo{number}{6} (\bibinfo{year}{2007}),
  \bibinfo{pages}{2769--2794}.
\newblock


\bibitem[\protect\citeauthoryear{Teney, Kafle, Shrestha, Abbasnejad, Kanan, and
  Hengel}{Teney et~al\mbox{.}}{2020}]%
        {teney2020value}
\bibfield{author}{\bibinfo{person}{Damien Teney}, \bibinfo{person}{Kushal
  Kafle}, \bibinfo{person}{Robik Shrestha}, \bibinfo{person}{Ehsan Abbasnejad},
  \bibinfo{person}{Christopher Kanan}, {and} \bibinfo{person}{Anton van~den
  Hengel}.} \bibinfo{year}{2020}\natexlab{}.
\newblock \showarticletitle{On the Value of Out-of-Distribution Testing: An
  Example of Goodhart's Law}. In \bibinfo{booktitle}{\emph{NeurIPS}}.
\newblock


\bibitem[\protect\citeauthoryear{Tran, Bourdev, Fergus, Torresani, and
  Paluri}{Tran et~al\mbox{.}}{2015}]%
        {tran2015learning}
\bibfield{author}{\bibinfo{person}{Du Tran}, \bibinfo{person}{Lubomir Bourdev},
  \bibinfo{person}{Rob Fergus}, \bibinfo{person}{Lorenzo Torresani}, {and}
  \bibinfo{person}{Manohar Paluri}.} \bibinfo{year}{2015}\natexlab{}.
\newblock \showarticletitle{Learning spatiotemporal features with 3d
  convolutional networks}. In \bibinfo{booktitle}{\emph{ICCV}}.
  \bibinfo{pages}{4489--4497}.
\newblock


\bibitem[\protect\citeauthoryear{Vapnik}{Vapnik}{1999}]%
        {vapnik1999overview}
\bibfield{author}{\bibinfo{person}{Vladimir~N Vapnik}.}
  \bibinfo{year}{1999}\natexlab{}.
\newblock \showarticletitle{An overview of statistical learning theory}.
\newblock \bibinfo{journal}{\emph{IEEE transactions on neural networks}}
  \bibinfo{volume}{10}, \bibinfo{number}{5} (\bibinfo{year}{1999}),
  \bibinfo{pages}{988--999}.
\newblock


\bibitem[\protect\citeauthoryear{Vaswani, Shazeer, Parmar, Uszkoreit, Jones,
  Gomez, Kaiser, and Polosukhin}{Vaswani et~al\mbox{.}}{2017}]%
        {vaswani2017attention}
\bibfield{author}{\bibinfo{person}{Ashish Vaswani}, \bibinfo{person}{Noam
  Shazeer}, \bibinfo{person}{Niki Parmar}, \bibinfo{person}{Jakob Uszkoreit},
  \bibinfo{person}{Llion Jones}, \bibinfo{person}{Aidan~N Gomez},
  \bibinfo{person}{{\L}ukasz Kaiser}, {and} \bibinfo{person}{Illia
  Polosukhin}.} \bibinfo{year}{2017}\natexlab{}.
\newblock \showarticletitle{Attention is all you need}. In
  \bibinfo{booktitle}{\emph{NeurIPS}}. \bibinfo{pages}{5998--6008}.
\newblock


\bibitem[\protect\citeauthoryear{Wang, Zha, Chen, Xiong, and Luo}{Wang
  et~al\mbox{.}}{2020d}]%
        {wang2020dual}
\bibfield{author}{\bibinfo{person}{Hao Wang}, \bibinfo{person}{Zheng-Jun Zha},
  \bibinfo{person}{Xuejin Chen}, \bibinfo{person}{Zhiwei Xiong}, {and}
  \bibinfo{person}{Jiebo Luo}.} \bibinfo{year}{2020}\natexlab{d}.
\newblock \showarticletitle{Dual Path Interaction Network for Video Moment
  Localization}. In \bibinfo{booktitle}{\emph{ACM MM}}.
  \bibinfo{pages}{4116--4124}.
\newblock


\bibitem[\protect\citeauthoryear{Wang, Ma, and Jiang}{Wang
  et~al\mbox{.}}{2020c}]%
        {wang2019temporally}
\bibfield{author}{\bibinfo{person}{Jingwen Wang}, \bibinfo{person}{Lin Ma},
  {and} \bibinfo{person}{Wenhao Jiang}.} \bibinfo{year}{2020}\natexlab{c}.
\newblock \showarticletitle{Temporally Grounding Language Queries in Videos by
  Contextual Boundary-aware Prediction}. In \bibinfo{booktitle}{\emph{AAAI}}.
\newblock


\bibitem[\protect\citeauthoryear{Wang, Huang, Zhang, and Sun}{Wang
  et~al\mbox{.}}{2020a}]%
        {wang2020visual}
\bibfield{author}{\bibinfo{person}{Tan Wang}, \bibinfo{person}{Jianqiang
  Huang}, \bibinfo{person}{Hanwang Zhang}, {and} \bibinfo{person}{Qianru Sun}.}
  \bibinfo{year}{2020}\natexlab{a}.
\newblock \showarticletitle{Visual commonsense r-cnn}. In
  \bibinfo{booktitle}{\emph{CVPR}}. \bibinfo{pages}{10760--10770}.
\newblock


\bibitem[\protect\citeauthoryear{Wang, Feng, Xiangnan, Zhang, and Chua}{Wang
  et~al\mbox{.}}{2021}]%
        {wang2021clickbait}
\bibfield{author}{\bibinfo{person}{Wenjie Wang}, \bibinfo{person}{Fuli Feng},
  \bibinfo{person}{He Xiangnan}, \bibinfo{person}{Hanwang Zhang}, {and}
  \bibinfo{person}{Tat-Seng Chua}.} \bibinfo{year}{2021}\natexlab{}.
\newblock \showarticletitle{Clicks can be Cheating: Counterfactual
  Recommendation for Mitigating Clickbait Issue}. In
  \bibinfo{booktitle}{\emph{SIGIR}}.
\newblock


\bibitem[\protect\citeauthoryear{Wang, Liang, Charlin, and Blei}{Wang
  et~al\mbox{.}}{2020b}]%
        {wang2020causal}
\bibfield{author}{\bibinfo{person}{Yixin Wang}, \bibinfo{person}{Dawen Liang},
  \bibinfo{person}{Laurent Charlin}, {and} \bibinfo{person}{David~M Blei}.}
  \bibinfo{year}{2020}\natexlab{b}.
\newblock \showarticletitle{Causal Inference for Recommender Systems}. In
  \bibinfo{booktitle}{\emph{Fourteenth ACM Conference on Recommender Systems}}.
  \bibinfo{pages}{426--431}.
\newblock


\bibitem[\protect\citeauthoryear{Xu, Ba, Kiros, Cho, Courville, Salakhudinov,
  Zemel, and Bengio}{Xu et~al\mbox{.}}{2015}]%
        {xu2015show}
\bibfield{author}{\bibinfo{person}{Kelvin Xu}, \bibinfo{person}{Jimmy Ba},
  \bibinfo{person}{Ryan Kiros}, \bibinfo{person}{Kyunghyun Cho},
  \bibinfo{person}{Aaron Courville}, \bibinfo{person}{Ruslan Salakhudinov},
  \bibinfo{person}{Rich Zemel}, {and} \bibinfo{person}{Yoshua Bengio}.}
  \bibinfo{year}{2015}\natexlab{}.
\newblock \showarticletitle{Show, attend and tell: Neural image caption
  generation with visual attention}. In \bibinfo{booktitle}{\emph{ICML}}.
  \bibinfo{pages}{2048--2057}.
\newblock


\bibitem[\protect\citeauthoryear{Yang, Dong, Cao, Wang, Wang, and Chua}{Yang
  et~al\mbox{.}}{2020a}]%
        {yang2020tree}
\bibfield{author}{\bibinfo{person}{Xun Yang}, \bibinfo{person}{Jianfeng Dong},
  \bibinfo{person}{Yixin Cao}, \bibinfo{person}{Xun Wang},
  \bibinfo{person}{Meng Wang}, {and} \bibinfo{person}{Tat-Seng Chua}.}
  \bibinfo{year}{2020}\natexlab{a}.
\newblock \showarticletitle{Tree-Augmented Cross-Modal Encoding for
  Complex-Query Video Retrieval}. In \bibinfo{booktitle}{\emph{SIGIR}}.
  \bibinfo{pages}{1339--1348}.
\newblock


\bibitem[\protect\citeauthoryear{Yang, Liu, Jian, Gao, and Wang}{Yang
  et~al\mbox{.}}{2020b}]%
        {yang2020weakly}
\bibfield{author}{\bibinfo{person}{Xun Yang}, \bibinfo{person}{Xueliang Liu},
  \bibinfo{person}{Meng Jian}, \bibinfo{person}{Xinjian Gao}, {and}
  \bibinfo{person}{Meng Wang}.} \bibinfo{year}{2020}\natexlab{b}.
\newblock \showarticletitle{Weakly-Supervised Video Object Grounding by
  Exploring Spatio-Temporal Contexts}. In \bibinfo{booktitle}{\emph{ACM MM}}.
  \bibinfo{pages}{1939--1947}.
\newblock


\bibitem[\protect\citeauthoryear{Yang, Wang, and Tao}{Yang
  et~al\mbox{.}}{2018}]%
        {yang2017person}
\bibfield{author}{\bibinfo{person}{Xun Yang}, \bibinfo{person}{Meng Wang},
  {and} \bibinfo{person}{Dacheng Tao}.} \bibinfo{year}{2018}\natexlab{}.
\newblock \showarticletitle{Person re-identification with metric learning using
  privileged information}.
\newblock \bibinfo{journal}{\emph{IEEE Transactions on Image Processing}}
  \bibinfo{volume}{27}, \bibinfo{number}{2} (\bibinfo{year}{2018}),
  \bibinfo{pages}{791--805}.
\newblock


\bibitem[\protect\citeauthoryear{Yap, Tan, and Pang}{Yap et~al\mbox{.}}{2007}]%
        {yap2007discovering}
\bibfield{author}{\bibinfo{person}{Ghim-Eng Yap}, \bibinfo{person}{Ah-Hwee
  Tan}, {and} \bibinfo{person}{Hwee-Hwa Pang}.}
  \bibinfo{year}{2007}\natexlab{}.
\newblock \showarticletitle{Discovering and exploiting causal dependencies for
  robust mobile context-aware recommenders}.
\newblock \bibinfo{journal}{\emph{IEEE Transactions on Knowledge and Data
  Engineering}} \bibinfo{volume}{19}, \bibinfo{number}{7}
  (\bibinfo{year}{2007}), \bibinfo{pages}{977--992}.
\newblock


\bibitem[\protect\citeauthoryear{Yuan, Ma, Wang, Liu, and Zhu}{Yuan
  et~al\mbox{.}}{2019a}]%
        {NIPS2019_8344}
\bibfield{author}{\bibinfo{person}{Yitian Yuan}, \bibinfo{person}{Lin Ma},
  \bibinfo{person}{Jingwen Wang}, \bibinfo{person}{Wei Liu}, {and}
  \bibinfo{person}{Wenwu Zhu}.} \bibinfo{year}{2019}\natexlab{a}.
\newblock \showarticletitle{Semantic Conditioned Dynamic Modulation for
  Temporal Sentence Grounding in Videos}.
\newblock In \bibinfo{booktitle}{\emph{NeurIPS}}. \bibinfo{pages}{536--546}.
\newblock


\bibitem[\protect\citeauthoryear{Yuan, Mei, and Zhu}{Yuan
  et~al\mbox{.}}{2019b}]%
        {yuan2019find}
\bibfield{author}{\bibinfo{person}{Yitian Yuan}, \bibinfo{person}{Tao Mei},
  {and} \bibinfo{person}{Wenwu Zhu}.} \bibinfo{year}{2019}\natexlab{b}.
\newblock \showarticletitle{To find where you talk: Temporal sentence
  localization in video with attention based location regression}. In
  \bibinfo{booktitle}{\emph{AAAI}}, Vol.~\bibinfo{volume}{33}.
  \bibinfo{pages}{9159--9166}.
\newblock


\bibitem[\protect\citeauthoryear{Yue, Zhang, Sun, and Hua}{Yue
  et~al\mbox{.}}{2020}]%
        {yue2020interventional}
\bibfield{author}{\bibinfo{person}{Zhongqi Yue}, \bibinfo{person}{Hanwang
  Zhang}, \bibinfo{person}{Qianru Sun}, {and} \bibinfo{person}{Xian-Sheng
  Hua}.} \bibinfo{year}{2020}\natexlab{}.
\newblock \showarticletitle{Interventional Few-Shot Learning}. In
  \bibinfo{booktitle}{\emph{NeurIPS}}.
\newblock


\bibitem[\protect\citeauthoryear{Zeng, Xu, Huang, Chen, Tan, and Gan}{Zeng
  et~al\mbox{.}}{2020}]%
        {zeng2020dense}
\bibfield{author}{\bibinfo{person}{Runhao Zeng}, \bibinfo{person}{Haoming Xu},
  \bibinfo{person}{Wenbing Huang}, \bibinfo{person}{Peihao Chen},
  \bibinfo{person}{Mingkui Tan}, {and} \bibinfo{person}{Chuang Gan}.}
  \bibinfo{year}{2020}\natexlab{}.
\newblock \showarticletitle{Dense regression network for video grounding}. In
  \bibinfo{booktitle}{\emph{CVPR}}.
\newblock


\bibitem[\protect\citeauthoryear{Zhang, Dai, Wang, Wang, and Davis}{Zhang
  et~al\mbox{.}}{2019a}]%
        {zhang2019man}
\bibfield{author}{\bibinfo{person}{Da Zhang}, \bibinfo{person}{Xiyang Dai},
  \bibinfo{person}{Xin Wang}, \bibinfo{person}{Yuan-Fang Wang}, {and}
  \bibinfo{person}{Larry~S Davis}.} \bibinfo{year}{2019}\natexlab{a}.
\newblock \showarticletitle{Man: Moment alignment network for natural language
  moment retrieval via iterative graph adjustment}. In
  \bibinfo{booktitle}{\emph{CVPR}}. \bibinfo{pages}{1247--1257}.
\newblock


\bibitem[\protect\citeauthoryear{Zhang, Zhang, Tang, Hua, and Sun}{Zhang
  et~al\mbox{.}}{2020c}]%
        {zhang2020causal}
\bibfield{author}{\bibinfo{person}{Dong Zhang}, \bibinfo{person}{Hanwang
  Zhang}, \bibinfo{person}{Jinhui Tang}, \bibinfo{person}{Xiansheng Hua}, {and}
  \bibinfo{person}{Qianru Sun}.} \bibinfo{year}{2020}\natexlab{c}.
\newblock \showarticletitle{Causal intervention for weakly-supervised semantic
  segmentation}. In \bibinfo{booktitle}{\emph{NeurIPS}}.
\newblock


\bibitem[\protect\citeauthoryear{Zhang, Sun, Jing, and Zhou}{Zhang
  et~al\mbox{.}}{2020b}]%
        {zhang-etal-2020-span}
\bibfield{author}{\bibinfo{person}{Hao Zhang}, \bibinfo{person}{Aixin Sun},
  \bibinfo{person}{Wei Jing}, {and} \bibinfo{person}{Joey~Tianyi Zhou}.}
  \bibinfo{year}{2020}\natexlab{b}.
\newblock \showarticletitle{Span-based Localizing Network for Natural Language
  Video Localization}. In \bibinfo{booktitle}{\emph{ACL}}.
  \bibinfo{address}{Online}, \bibinfo{pages}{6543--6554}.
\newblock


\bibitem[\protect\citeauthoryear{Zhang, Peng, Fu, and Luo}{Zhang
  et~al\mbox{.}}{2020a}]%
        {zhang2019learning}
\bibfield{author}{\bibinfo{person}{Songyang Zhang}, \bibinfo{person}{Houwen
  Peng}, \bibinfo{person}{Jianlong Fu}, {and} \bibinfo{person}{Jiebo Luo}.}
  \bibinfo{year}{2020}\natexlab{a}.
\newblock \showarticletitle{Learning 2D Temporal Adjacent Networks for Moment
  Localization with Natural Language}. In \bibinfo{booktitle}{\emph{AAAI}}.
\newblock


\bibitem[\protect\citeauthoryear{Zhang, Lin, Zhao, and Xiao}{Zhang
  et~al\mbox{.}}{2019b}]%
        {zhang2019cross}
\bibfield{author}{\bibinfo{person}{Zhu Zhang}, \bibinfo{person}{Zhijie Lin},
  \bibinfo{person}{Zhou Zhao}, {and} \bibinfo{person}{Zhenxin Xiao}.}
  \bibinfo{year}{2019}\natexlab{b}.
\newblock \showarticletitle{Cross-modal interaction networks for query-based
  moment retrieval in videos}. In \bibinfo{booktitle}{\emph{SIGIR}}.
  \bibinfo{pages}{655--664}.
\newblock


\end{thebibliography}

\end{document}